\documentclass{article} 
\usepackage{iclr2026_conference,times}

\usepackage{amsmath,amsfonts,bm}

\def\eqref#1{equation~\ref{#1}}

\def\1{\bm{1}}

\DeclareMathAlphabet{\mathsfit}{\encodingdefault}{\sfdefault}{m}{sl}
\SetMathAlphabet{\mathsfit}{bold}{\encodingdefault}{\sfdefault}{bx}{n}

\usepackage{hyperref}
\usepackage{url}
\usepackage{marginnote}

\usepackage{graphicx}

\usepackage{blindtext}
\usepackage{comment}
\usepackage{xcolor}
\usepackage{marginnote}
\usepackage{amsmath}

\usepackage{booktabs}
\usepackage{colortbl}      
\usepackage{pifont}        
\usepackage{fontawesome5}  
\usepackage{wrapfig}
\usepackage{caption}

\newcommand{\cmark}{\ding{51}}
\newcommand{\xmark}{\ding{55}}

\usepackage{listings}
\usepackage[most]{tcolorbox}

\definecolor{PromptBG}{RGB}{240,248,255}    
\definecolor{PromptFrame}{RGB}{60,120,200}  
\definecolor{PromptTitleBG}{RGB}{220,235,250} 

\lstdefinestyle{prompt}{
  basicstyle=\ttfamily\small,
  breaklines=true,
  breakatwhitespace=true,
  breakindent=0pt,
  frame=single,
  framerule=0.4pt,
  showstringspaces=false,
  columns=fullflexible,
  keepspaces=true
}

\newtcolorbox{PromptBox}[1][]{
  enhanced,
  breakable,
  colback=PromptBG,        
  colframe=PromptFrame,    
  coltitle=black,
  boxrule=0.8pt,
  arc=3pt,
  left=7pt,right=7pt,top=7pt,bottom=7pt,
  fonttitle=\bfseries,
  attach boxed title to top left={yshift=-2pt, xshift=6pt},
  boxed title style={
    colback=PromptTitleBG, 
    colframe=PromptFrame,
    coltitle=black,
    boxrule=0.8pt,
    arc=3pt,
    left=6pt,right=6pt,top=2pt,bottom=2pt
  },
  title={#1}
}

\definecolor{DSPyBG}{RGB}{235,247,246}
\definecolor{DSPyFrame}{RGB}{0,128,128}
\definecolor{DSPyTitleBG}{RGB}{214,241,238}
\definecolor{DSPyInstrBG}{RGB}{246,250,250}

\lstdefinestyle{dspy-instr}{
  basicstyle=\ttfamily\small,
  breaklines=true,
  breakatwhitespace=true,
  frame=single,
  framerule=0.4pt,
  showstringspaces=false,
  columns=fullflexible,
  keepspaces=true,
  backgroundcolor=\color{DSPyInstrBG}
}

\newtcolorbox{DSPyBox}[1][]{
  enhanced,
  breakable,
  colback=DSPyBG,
  colframe=DSPyFrame,
  boxrule=0.9pt,
  arc=3pt,
  left=8pt,right=8pt,top=8pt,bottom=8pt,
  coltitle=black,
  fonttitle=\bfseries\ttfamily,
  attach boxed title to top left={yshift=-2pt, xshift=6pt},
  boxed title style={
    colback=DSPyTitleBG,
    colframe=DSPyFrame,
    boxrule=0.9pt,
    arc=3pt,
    left=6pt,right=6pt,top=2pt,bottom=2pt
  },
  title={#1}
}

\newcommand{\code}[1]{\texttt{\detokenize{#1}}}

\newcolumntype{L}[1]{>{\raggedright\arraybackslash}p{#1}}
\newenvironment{DSPyFields}
  {\begin{tabular*}{\linewidth}{@{\extracolsep{\fill}} L{0.24\linewidth} L{0.28\linewidth} L{0.43\linewidth} }
   \toprule \textbf{Field} & \textbf{Type} & \textbf{Description} \\\midrule}
  {\bottomrule\end{tabular*}}

\newcommand{\DSPyField}[3]{\code{#1} & \code{#2} & #3\\}

\definecolor{AutoBG}{RGB}{246, 242, 252}      
\definecolor{AutoFrame}{RGB}{120, 80, 180}    
\definecolor{AutoTitleBG}{RGB}{234, 226, 248} 
\definecolor{AutoDescBG}{RGB}{252, 250, 255}  

\newtcolorbox{AutoBox}[1][]{
  enhanced,
  breakable,
  colback=AutoBG,
  colframe=AutoFrame,
  boxrule=0.9pt,
  arc=3pt,
  left=8pt,right=8pt,top=8pt,bottom=10pt,
  coltitle=black,          
  fonttitle=\bfseries,
  attach boxed title to top left={yshift=-2pt, xshift=6pt},
  boxed title style={
    colback=AutoTitleBG,
    colframe=AutoFrame,
    coltitle=black,        
    boxrule=0.9pt,
    arc=3pt,
    left=6pt,right=6pt,top=2pt,bottom=2pt
  },
  title={#1}
}

\newtcolorbox{AutoDesc}[1][]{
  enhanced, breakable, 
  colback=AutoDescBG,
  colframe=AutoFrame!40,
  coltitle=black,          
  fonttitle=\bfseries,
  boxrule=0.5pt,
  arc=2pt,
  left=6pt,right=6pt,top=6pt,bottom=6pt,
  title={#1}
}

\newcolumntype{L}[1]{>{\raggedright\arraybackslash}p{#1}}
\newcolumntype{R}[1]{>{\raggedleft\arraybackslash}p{#1}}

\title{AutoMetrics: Approximate Human Judgments with Automatically Generated Evaluators}

\author{
Michael J. Ryan\textsuperscript{$\spadesuit$}, 
Yanzhe Zhang\textsuperscript{$\spadesuit$}, 
Amol Salunkhe\textsuperscript{$\clubsuit$}, 
Yi Chu\textsuperscript{$\clubsuit$}, 
Di Xu\textsuperscript{$\clubsuit$}, 
Diyi Yang\textsuperscript{$\spadesuit$} \\
\textsuperscript{$\spadesuit$}Stanford University \quad
\textsuperscript{$\clubsuit$}American Express \\
\texttt{michaeljryan@stanford.edu}
}

\iclrfinalcopy 
\begin{document}

\maketitle
\begingroup
\renewcommand{\thefootnote}{}%
\footnotetext{This paper reflects the academic work of the authors and does not represent or constitute the views, policies, positions, or practices of American Express or its affiliates.}%
\endgroup
\setcounter{footnote}{0}

\begin{abstract}
Evaluating user-facing AI applications remains a central challenge, especially in open-ended domains such as travel planning, clinical note generation, or dialogue. The gold standard is user feedback (e.g., thumbs up/down) or behavioral signals (e.g., retention), but these are often scarce in prototypes and research projects, or too-slow to use for system optimization. We present \textbf{AutoMetrics}, a framework for synthesizing evaluation metrics under low-data constraints. AutoMetrics combines retrieval from \textbf{MetricBank}, a collection of 48 metrics we curate, with automatically generated LLM-as-a-Judge criteria informed by lightweight human feedback. These metrics are composed via regression to maximize correlation with human signal.  AutoMetrics takes you from expensive measures to interpretable automatic metrics. Across 5 diverse tasks, AutoMetrics improves Kendall correlation with human ratings by up to 33.4\% over LLM-as-a-Judge while requiring fewer than 100 feedback points. We show that AutoMetrics can be used as a proxy reward to equal effect as a verifiable reward. We release the full AutoMetrics toolkit and MetricBank to accelerate adaptive evaluation of LLM applications.
\end{abstract}

\begin{figure}[h]
    \centering
    \includegraphics[width=1.0\linewidth]{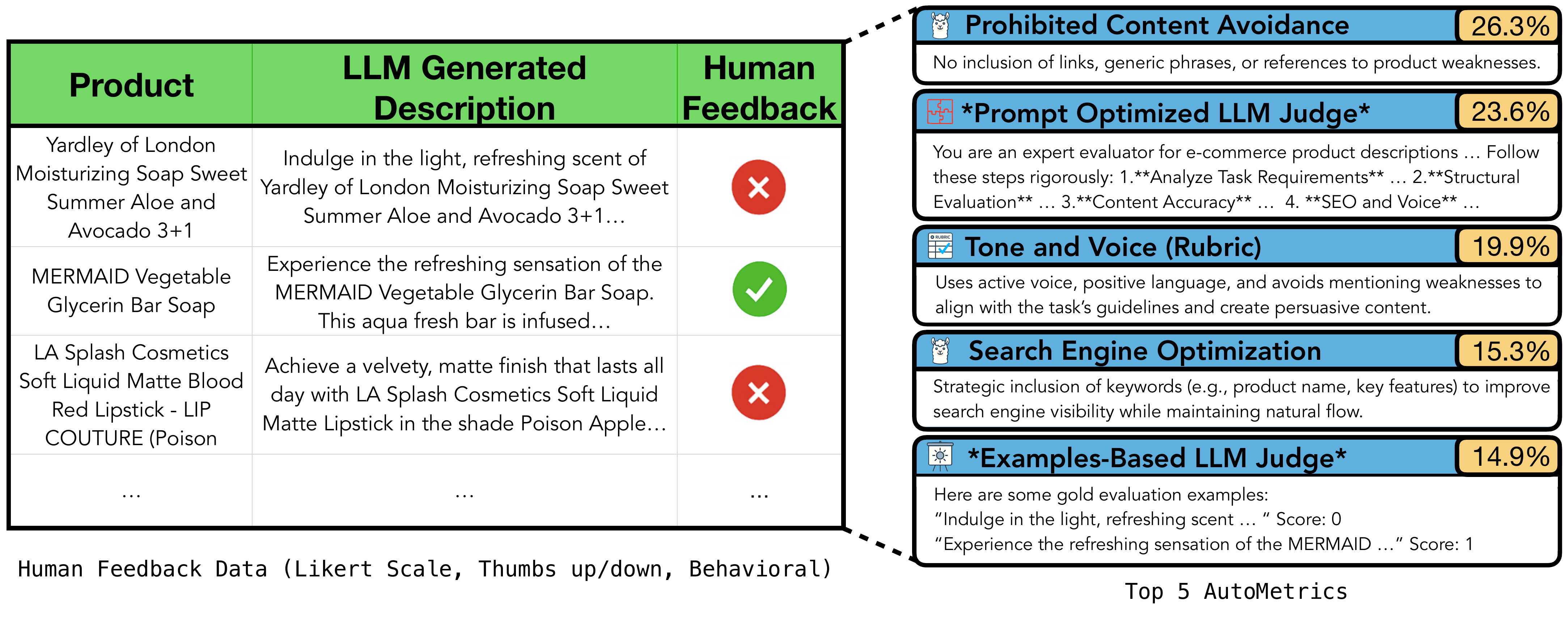}
    \caption{AutoMetrics takes you from expensive measures to interpretable automatic metrics.  Here AutoMetrics generates useful metrics for evaluating LLM written product descriptions from user reviews from EvalGen \citep{10.1145/3654777.3676450}.  Percentages indicate relative importance of each metric derived from regression coefficients.}
    \label{fig:motivation}
\end{figure}

\section{Introduction}
\label{sec:introduction}



Modern AI systems now demonstrate massively multitask capabilities imparted through extensive pretraining \citep{radford2019language, brown2020languagemodelsfewshotlearners}. Practitioners can rapidly prototype new AI-enabled tasks -- from travel planning to code completion -- at a pace much faster than the community can craft domain specific metrics \citep{papineni-etal-2002-bleu, lin-2004-rouge, xu-etal-2016-optimizing}. This new era, in which large language models can be adapted to virtually any domain, places mounting pressure on evaluation practices. A divide is growing between tasks with easily verifiable rewards, such as math \citep{glazer2024frontiermathbenchmarkevaluatingadvanced, shao2024deepseekmathpushinglimitsmathematical} and coding \citep{chen2021evaluatinglargelanguagemodels},  while subjective and open-ended tasks such as writing \citep{gurung2025learningreasonlongformstory} remain difficult to measure. For these tasks, human evaluation remains the gold standard \citep{10.1145/3654777.3676450, chiang2024chatbotarenaopenplatform}.

Unfortunately, human evaluation is costly, slow, and not scalable for every prototype or user population. Reward models offer an alternative \citep{mnih_human-level_2015, 10.5555/3294996.3295184}, but they typically require thousands of labels. The common alternative is rubric-based LLM-as-a-Judge methods \citep{alpaca_eval, zheng2023judging, liu2024enhancing}, which rely on the assumption that system behavior is clearly defined and are not guaranteed to follow given rubrics strictly  \citep{tripathi2025pairwise}. In reality, practitioners typically have access only to non-descriptive human signals (e.g., thumbs up/thumbs down collected from users). In this setting, the problem is not only formulating the rubric, but also discovering the underlying criteria that matter.




This highlights the need for \textbf{dynamic, task-specific metric learning}. Instead of relying exclusively on human judgment or fixed rubrics, evaluation itself must become adaptive. Current efforts have emphasized making LLMs better evaluators of task-specific criteria \citep{liu2024enhancing, kim-etal-2025-biggen, anugraha2025r3robustrubricagnosticreward} or leveraging rubrics to optimize LLMs \citep{gunjal2025rubricsrewardsreinforcementlearning, viswanathan2025checklistsbetterrewardmodels} but comparatively little work has focused on automatically generating the rubrics and criteria to be adaptively aligned with human judgment \citep{10.1145/3664646.3664778, ryan-etal-2025-synthesizeme, dunlap2025vibecheckdiscoverquantifyqualitative}.
Such adaptive evaluation is essential not only for easily assessing new tasks but also for optimizing evaluated systems based on real-time user feedback.


We introduce \textbf{AutoMetrics}, a method for dynamic metric induction that turns sparse, non-descriptive human feedback into actionable and interpretable evaluators (Figure~\ref{fig:motivation}). Starting from a task description and fewer than 100 human signals, AutoMetrics synthesizes candidate criteria, retrieves and adapts existing metrics, and composes them through regression into predictive measures of quality. Beyond simply identifying criteria, \textbf{AutoMetrics grounds and weighs them}, producing metrics that are both predictive and interpretable. This approach achieves up to \textbf{33.4\% higher Kendall correlation} with human judgments than LLM-as-a-Judge baselines (\S\ref{sec:experiments}), is \textbf{data-efficient} only requiring $\sim$80 feedback points (\S\ref{subsec:data_efficiency}), and even \textbf{matches verifiable rewards} when optimizing downstream AI systems (\S\ref{sec:case_study}). Beyond accuracy, \textbf{AutoMetrics reveals actionable insights into what users value}. We release AutoMetrics as an open-source toolkit\footnote{\url{https://github.com/SALT-NLP/autometrics}}, offering the community a powerful new way to evaluate and optimize AI applications at the speed of modern development.

\section{Related Work}
\label{sec:related_work}


\paragraph{Metric Collections}
Prior work has organized collections of metrics primarily for the ease of use on the part of the practitioner.  When already using a library such as PyTorch \citep{10.5555/3454287.3455008} or Huggingface \citep{wolf-etal-2020-transformers} it's simple to utilize TorchMetrics \citep{torchmetrics} or HuggingFace \texttt{lighteval} \citep{lighteval}.  Scikit Learn Metrics \citep{scikit-learn} and NLTK metrics \citep{bird-loper-2004-nltk} were created with the same intentions.  All text-generation metrics covered by these collections are also contained in our MetricBank collection.
Beyond integrating with existing open source libraries, some metric collections are part of ML observability frameworks like Evidently \citep{evidentlyai2025}, Galileo \citep{galileo2025promptquality}, Scorecard.io \citep{doe2024scorecardai}, and DeepEval \citep{confidentai2025deepeval}. Most metrics are tightly coupled with their observability platform, although Evidently and DeepEval offer open-source versions.  While DeepEval offers a metric recommendation feature, it is based on a predefined decision tree of questions like ``\emph{Does your LLM application use Retrieval-Augmented Generation (RAG)?}" and ``\emph{Is LLM safety a priority for you?}".  Most similar to our work is the MetaMetrics collection \citep{winata2025metametrics}, which computes a regression over multiple task-specific metrics for tasks like image captioning and summarization to select the best combination of metrics.  We compare our approach with MetaMetrics in Section \ref{sec:experiments} and find that our core thesis of adaptive metric generation is critical for evaluation in the low-data, novel task settings of interest.

\paragraph{LLM Based Evaluation}
LLM-as-a-Judge \citep{zheng2023judging} evaluation is increasingly popular with the frequent improvement of LLM capabilities. Several works devise task-specific prompts to enable LLM-based evaluation for storytelling \citep{chiang-lee-2023-large}, summarization \citep{wang-etal-2023-chatgpt, hada-etal-2024-metal, wu2023largelanguagemodelsdiverse}, dialogue \citep{lin-chen-2023-llm, fu-etal-2024-gptscore}, knowledge \citep{bai2023benchmarking}, translation \citep{kocmi-federmann-2023-large}, and more \citep{brake-schaaf-2024-comparing}.  
Another promising direction is devising frameworks and general methods for making LLM-as-a-Judge more reliable. G-Eval \citep{liu-etal-2023-g} proposes breaking LLM evaluation into a step-by-step chain of thought and taking a weighted sum over the log probabilities of generating different scores.  ChatEval \citep{chan2024chateval} simulates multiple perspectives by evaluating through multi-agent debate.  SPADE \citep{10.14778/3685800.3685835} generates assertions for LLMs to verify based on labeled good and bad examples.  VERDICT \citep{kalra2025verdictlibraryscalingjudgetime} introduces judge-time scaling by decomposing judgments into composable units of reasoning, verification, debate, and aggregation steps.  Though we take inspiration from many of these frameworks, the most directly similar to our LLM-as-a-Judge steps in the AutoMetrics pipeline are DnA-Eval \citep{li-etal-2025-dna} and EvalGen \citep{10.1145/3654777.3676450}.  DnA-Eval \citep{li-etal-2025-dna} decomposes evaluation into rubric criteria and aggregates the results across the criteria.  EvalGen \citep{10.1145/3654777.3676450} elicits limited human feedback on generated outputs, proposes criteria for evaluation based on this feedback, and iteratively refines the criteria with a human-in-the-loop and LLM.
\section{The AutoMetrics Method}
\label{sec:auto_metric_pipeline}

\begin{figure*}
    \centering
    \includegraphics[width=1.0\textwidth]{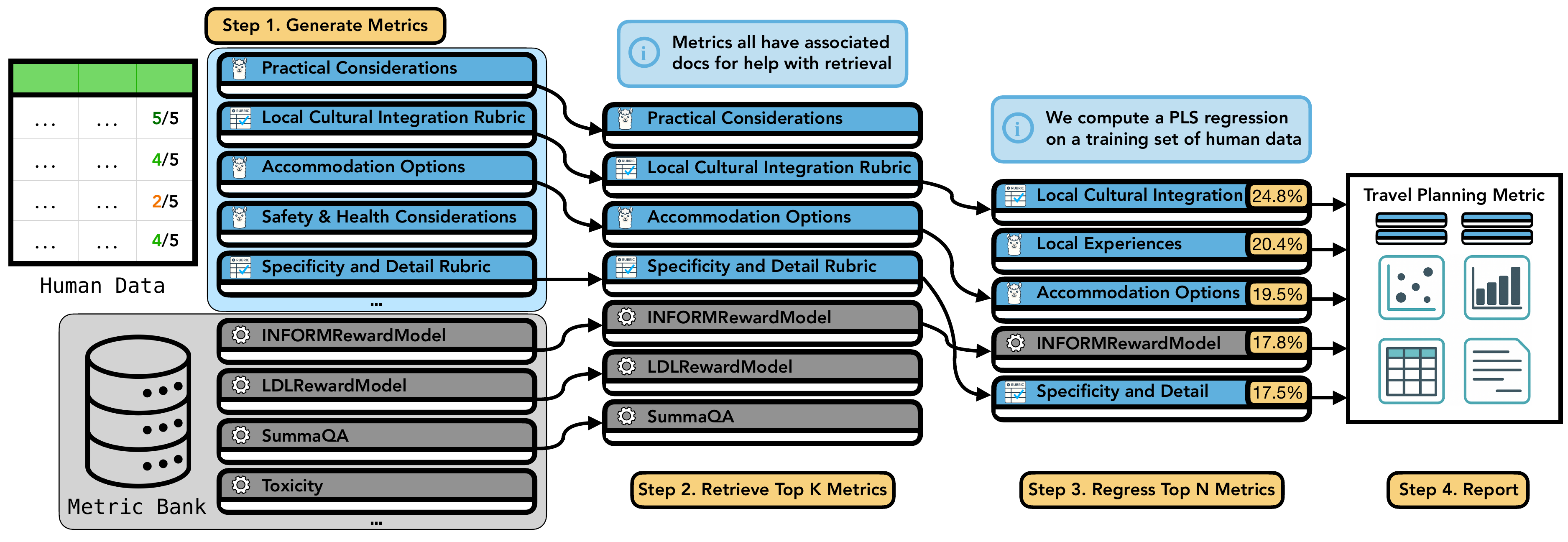}
    \caption{\textbf{AutoMetrics} comprises four steps. (1) \emph{Generate}: create task-specific candidate metrics (Single criteria, Rubric, Examples, MIPROv2). (2) \emph{Retrieve}: from the generated candidates plus MetricBank, use ColBERT to prefilter to $k'$ metric cards and an LLM to select the final $k$. (3) \emph{Regress}: fit a PLS model on the training set to weight and select metrics that predict human judgments. (4) \emph{Report}: produce a writeup with weights and correlations and details to guide adoption.}
    \label{fig:autometrics}
\end{figure*}

The purpose of AutoMetrics is to produce metrics for subjective and novel AI-enabled tasks. Our goal is to induce metrics that correlate strongly with human judgments while requiring minimal data collection. To accomplish this, we present a general pipeline with four stages: (1) generate, (2) retrieve, (3) regress, and (4) report. These steps are visualized in Figure~\ref{fig:autometrics}. Each stage involves design choices among several alternatives, which we empirically validate (\S\ref{subsec:ablations}).

\subsection{Metric Production}

\paragraph{Generate}
\label{subsec:generate}
For sufficiently novel settings, generating criteria for LLM-as-a-Judge evaluation is essential. Broad coverage of evaluation criteria allows us to later filter down to what matters most. Accordingly, our default configuration generates \textbf{10 Single Criterion} LLM Judge metrics, \textbf{5 Rubric} LLM-Judge metrics, \textbf{1 Example}-based optimized LLM-Judge metric (fewshot), and \textbf{1 Prompt-Optimized} LLM-Judge metric per run of AutoMetrics\footnote{Design details and ablations are in Appendix~\ref{appendix:subsec:generate}; we validate these choices across nearly 30 settings.}.  Optimized metrics require more LLM calls/tokens to produce, while criteria and rubrics are relatively inexpensive. Unless otherwise specified, we use this configuration throughout the paper. Empirically, we find this mix of generated metrics generalizes across diverse domains and tasks. For each metric, we also generate a Metric Card documenting its description, intended use, implementation details, and limitations (Appendix~\ref{sec:metric_cards}).

\paragraph{Retrieve}
\label{subsec:retrieve}
In addition to generated metrics, we leverage our MetricBank: a collection of 48 metrics (Appendix Table~\ref{tab:metrics}) drawn from the NLP literature, each implemented and documented with a Metric Card. Directly evaluating all metrics would be prohibitively expensive, so we instead use retrieval as a filtering step. We treat Metric Cards as documents, and use a description of the evaluation setting or task as the search query. Retrieval is performed using a hybrid \textbf{ColBERT + LLM} approach,\footnote{We ablate the selection algorithm in Appendix~\ref{appendix:subsec:retrieve}.} narrowing the candidate pool to metrics most relevant to the task at hand.

\paragraph{Regress}
\label{subsec:regression}
The filtered pool of candidate metrics must still be combined into a predictive signal for human judgment. We normalize all metric scores to their z-scores and fit a \textbf{Partial Least Squares (PLS)} regression model.  
Intuitively, PLS projects the metric space onto the direction most predictive of human labels, then regresses labels along that axis. We choose PLS regression because it works well under the constraints of our setting that: (1) the number of predictors (metrics) may be comparable to or larger than the number of observations (data points), and (2) the predictors are often highly correlated.
Concretely, with a single latent component, PLS finds a unit weight vector $w^\star \in \mathbb{R}^d$ that maximizes
\[
w^\star \;=\; \arg\max_{\lVert w\rVert_2 = 1} \; \operatorname{cov}(X w,\, y)^2,
\]
where $X$ is the matrix of normalized metric scores and $y$ is the vector of human labels. The latent score is $t = X w^\star$, and PLS then regresses the human labels on this latent score, yielding predictions $\hat{y} = t \beta$ with coefficient $\beta = \tfrac{t^\top y}{t^\top t}$.

We apply this procedure in two stages. In the first stage, we fit PLS using all candidate metrics and rank them by the magnitude of their weights in $w^\star$. We then select the top $n$ metrics according to this ranking. In the second stage, we refit PLS on this reduced set of $n$ metrics to obtain a new projection $t$ and corresponding predictions $\hat{y}$.
As a final step, we remove negatively correlated LLM-generated metrics, as they are designed to target positive correlation. We don't apply this to existing measures (e.g., length can negatively correlate with conciseness).

\subsection{Metric Evaluation}
\label{subsec:evaluation-principles}

To evaluate the quality of induced metrics, we draw on concepts of measurement validity from research \citep{borsboom2004concept} and testing \citep{AERA_APA_NCME_2014}. We focus on three forms: ``Content Validity", ``Criterion Validity", and ``Construct Validity".

\textbf{Content Validity} asks whether a metric represents the construct it is intended to measure.  Although direct quantification is difficult, we encourage transparency by releasing metric reports. Because our generated metrics rely on LLM judges, we also expose the reasoning traces of the judge LLM, allowing users to inspect whether assessments appear justified. These traces can further aid system optimization with AutoMetrics (\S \ref{sec:case_study}).

\textbf{Criterion Validity} Criterion validity measures correlation with a reference standard. In NLP, correlation with human labels has been the most widely used criterion \citep{banerjee-lavie-2005-meteor, xu-etal-2016-optimizing, gehrmann-etal-2021-gem}. We assess criterion validity by comparing AutoMetrics to ground-truth human labels. We report Kendall’s $\tau$, which makes no distributional assumptions and simply checks whether the rank order induced by a metric matches that of human judgments. This provides a conservative estimate compared to Spearman’s $\rho$ or Pearson’s $r$.

\textbf{Construct Validity} measures whether a metric captures an underlying abstract concept, such as “quality.” Both human judgments and AutoMetrics attempt to approximate ``quality''. We draw from convergent–discriminant validity \citep{campbell1959convergent} and operationalize construct validity as robustness. A useful metric should penalize quality degradations (sensitivity) while remaining stable under equivalent-quality variation.
In order to quantify convergent-discriminant validity, we introduce two measurements: \textbf{Sensitivity} and \textbf{Stability}. To construct test cases, we use an LLM to generate strategies for degrading outputs on a given dataset, and apply these to produce \emph{worse-quality perturbations}. In contrast, \emph{same-quality perturbations} are produced from a fixed set of hand-crafted transformations—such as rephrasing, reordering, synonym replacement, or stylistic edits—that are designed to preserve the target evaluation dimension. Prompts are provided in Appendix~\ref{appendix:prompts}.

\begin{itemize}
\item \textbf{Sensitivity} measures whether a metric assigns lower scores to degraded outputs. Let $s^{(i)}_{\text{orig}}$ and $s^{(i)}_{\text{worse}}$ denote the normalized scores for the original and worse-quality perturbed outputs of sample $i$ from a dataset of size $|N|$. Sensitivity is defined as:
\[
\text{Sensitivity} = \tfrac{1}{N} \sum_{i=1}^N \mathbf{1}\!\left[s^{(i)}_{\text{worse}} < s^{(i)}_{\text{orig}}\right]
\]

\item \textbf{Stability} measures whether a metric produces consistent scores when quality should be preserved. Let $s^{(i)}_{\text{same}}$ be the normalized score for a same-quality perturbation of sample $i$ from a dataset of size $|N|$. Stability is defined as:
\[
\text{Stability} = 1 - \tfrac{1}{N} \sum_{i=1}^N \big| s^{(i)}_{\text{orig}} - s^{(i)}_{\text{same}} \big|.
\]
 
\end{itemize}

High sensitivity indicates strong penalization of degraded outputs, while high stability indicates invariance to irrelevant variation. Both are desirable, and together they provide a general-purpose lens for evaluating how well a metric generalizes.

\section{Experiments and Evaluations: Showing AutoMetrics are Valid}
\label{sec:experiments}

For our experiments, we focus on showing that our AutoMetrics are valid across many tasks/domains and that they correlate better with human judgements than competitive baselines.  We showcase AutoMetrics have high Criterion Validity and Construct Validity across several tasks.

\subsection{Tasks}
\label{subsec:tasks}

\begin{table*}[h!]
  \centering
  \small
  \setlength{\tabcolsep}{6pt}
  \renewcommand{\arraystretch}{1.1}
  \resizebox{\textwidth}{!}{%
    \begin{tabular}{l l c c c c c}
      \toprule
      \rowcolor{gray!30}
      \textbf{Dataset (Citation)} 
        & \textbf{Task} 
        & \textbf{Domain} 
        & \textbf{\# Data} 
        & \textbf{Feedback} 
        & \textbf{\# Eval Dim} 
        & \textbf{Refs} \\
      \midrule
      \rowcolor{gray!10}
      \multicolumn{7}{l}{\textit{\textbf{In-Distribution Tasks}: some metrics in our bank were designed to directly evaluate these tasks.}}\\
      \midrule
      SimpEval \citep{maddela-etal-2023-lens}  
        & Simplification  
        & \faIcon{book-reader}  
        & 360  
        & 1--100 Likert  
        & 1
        & \cmark \\
      HelpSteer2 \citep{wang2024helpsteer}  
        & Dialogue  
        & \faIcon{comments}  
        & 20,324  
        & 1--5 Likert  
        & 5  
        & \xmark \\
      \midrule
      \rowcolor{gray!10}
      \multicolumn{7}{l}{\textit{\textbf{Out-of-Distribution Tasks}: no metric is specifically designed for these -- tests generalization and metric generation.}}\\
      \midrule
      EvalGen \citep{10.1145/3654777.3676450}  
        & Product description  
        & \faIcon{file-invoice}  
        & 100  
        & Binary  
        & 1  
        & \xmark \\
      RealHumanEval \citep{mozannar2025the}  
        & Code completion  
        & \faIcon{code}  
        & 5,204  
        & Behavioral  
        & 1  
        & \xmark \\
      Co-Gym \citep{shao2025collaborativegymframeworkenabling}  
        & Travel planning  
        & \faIcon{route}  
        & 72  
        & 1--5 Likert  
        & 3  
        & \xmark \\
      \bottomrule
    \end{tabular}%
  }
  \caption{Overview of tasks.  
    \textbf{Icons:}  
    \faIcon{code} Code Generation;  
    \faIcon{file-invoice} Data-to-Text Generation;  
    \faIcon{comments} Dialogue/Chat;  
    \faIcon{book-reader} Education/Readability;  
    \faIcon{route} Travel Planning.
    }
  \label{tab:autometrics_tasks}
\end{table*}

In order to evaluate our AutoMetrics method, we collect two types of tasks: \textit{In-Distribution Tasks}, which are tasks where some of the metrics in our Metric Bank were designed to directly evaluate the task, and \textit{Out-of-Distribution Tasks}, which are tasks where no metric in particular was designed to assess the task.  All of our tasks utilize human feedback for evaluation, encompassing behavioral feedback, binary feedback (thumbs up/down), and Likert scale feedback, which is already collected as part of the dataset.  We introduce all tasks in Table~\ref{tab:autometrics_tasks}.  In our main tables we present results for five datasets and a single evaluation dimension from each: \textbf{SimpEval} \citep{maddela-etal-2023-lens} (sentence simplification score 1--100), \textbf{HelpSteer2} \citep{wang2024helpsteer} (Chatbot helpfulness 1--5), \textbf{EvalGen} \citep{10.1145/3654777.3676450} (Product Review Thumbs Up/Down), \textbf{RealHumanEval} \citep{mozannar2025the} (accepted or rejected code edit), \textbf{CoGym} \citep{shao2025collaborativegymframeworkenabling} (travel plan outcome rating 1--5).  We report evaluations on more settings in the Appendix results.

\subsection{Baselines}
\label{subsec:baselines}

We include the following baselines: \textbf{Best Existing Metric}, where we run all 48 metrics (or 19 metrics for reference-free tasks), record their Kendall correlation on the validation set, and select the best metric to use for the task based on the validation correlation. \textbf{MetaMetrics}, where we take all the metrics from the MetaMetrics paper and compute an XGBoost Regression on the metrics on the trainset \citep{winata2025metametrics}.
\textbf{Finetuned LLM} refers to training a \texttt{ModernBERT-large} \citep{warner2024smarter} to predict the human annotation. We implement it by training LoRA adapters \citep{hu2021lora} with rank $= 16$ on all the attention, dense layers, and regression head, using a learning rate of $5e-5$ and a batch size of 16 for three epochs over the training data.  For the \textbf{LLM-Judge} baseline, we use the original human annotation prompt for each task and provide it to an LLM.  We include all of these prompts in Appendix~\ref{appendix:prompts}.  \textbf{DnA-Eval} \citep{li-etal-2025-dna} involves using an LLM to generate three dimensions where a user request may benefit from evaluation, along with weights for how to aggregate these dimensions. Then each of those dimensions is scored with an LLM-as-a-Judge, and finally aggregated based on the LLM-generated weights.


%

\subsection{Criterion Validity (Correlation)}
\label{subsec:correlation}

We report Kendall's $\tau$ of all methods with GPT-4o-mini and Qwen-3-32B Reasoning in Table~\ref{tab:main_results}.

\begin{table*}[h]
\centering
\setlength\tabcolsep{4pt}
\resizebox{\textwidth}{!}{
\begin{tabular}{l | cc|ccc}
\toprule
\rowcolor[gray]{0.9} & \multicolumn{2}{c|}{\textbf{In-Distribution}} & \multicolumn{3}{c}{\textbf{Out-of-Distribution}} \\
\rowcolor[gray]{0.9} Method & SimpEval & HelpSteer2 & EvalGen & RealHumanEval & CoGym \\
\midrule
\rowcolor[gray]{0.85} \multicolumn{6}{l}{\textbf{Model Agnostic}}\\
 Best Existing Metric & 0.246 \scriptsize$\pm$ 0.00 & 0.327 \scriptsize$\pm$ 0.00 & 0.193 \scriptsize$\pm$ 0.00 & 0.138 \scriptsize$\pm$ 0.00 & 0.074 \scriptsize$\pm$ 0.00 \\
 MetaMetrics \citep{winata2025metametrics} & 0.127 \scriptsize$\pm$ 0.01 & 0.204 \scriptsize$\pm$ 0.00 & -0.214 \scriptsize$\pm$ 0.01 & 0.025 \scriptsize$\pm$ 0.01 & -0.119 \scriptsize$\pm$ 0.02 \\
 Finetuned LLM & 0.076 \scriptsize$\pm$ 0.08 & 0.039 \scriptsize$\pm$ 0.03 & 0.054 \scriptsize$\pm$ 0.05 & 0.049 \scriptsize$\pm$ 0.06 & 0.223 \scriptsize$\pm$ 0.20 \\
\midrule
\rowcolor[gray]{0.85} \multicolumn{6}{l}{\textbf{GPT-4o-mini Backbone}}\\
 LLM-Judge & 0.272 \scriptsize$\pm$ 0.02 & 0.259 \scriptsize$\pm$ 0.01 & 0.161 \scriptsize$\pm$ 0.14 & 0.069 \scriptsize$\pm$ 0.01 & \textbf{0.199} \scriptsize$\pm$ 0.13 \\
 DnA Eval \citep{li-etal-2025-dna} & 0.234 \scriptsize$\pm$ 0.03 & 0.255 \scriptsize$\pm$ 0.02 & 0.174 \scriptsize$\pm$ 0.16 & 0.152 \scriptsize$\pm$ 0.01 & 0.185 \scriptsize$\pm$ 0.10 \\
 AutoMetrics (Ours) & \textbf{0.321} \scriptsize$\pm$ 0.04 & \textbf{0.324} \scriptsize$\pm$ 0.01 & \textbf{0.334} \scriptsize$\pm$ 0.06 & \textbf{0.160} \scriptsize$\pm$ 0.00 & -0.034 \scriptsize$\pm$ 0.17 \\
\midrule
\rowcolor[gray]{0.85} \multicolumn{6}{l}{\textbf{Qwen-3-32B Backbone}}\\
 LLM-Judge & 0.294 \scriptsize$\pm$ 0.04 & 0.334 \scriptsize$\pm$ 0.02 & 0.272 \scriptsize$\pm$ 0.13 & 0.025 \scriptsize$\pm$ 0.01 & 0.276 \scriptsize$\pm$ 0.19 \\
 DnA Eval \citep{li-etal-2025-dna} & 0.042 \scriptsize$\pm$ 0.04 & 0.260 \scriptsize$\pm$ 0.02 & 0.232 \scriptsize$\pm$ 0.19 & 0.071 \scriptsize$\pm$ 0.15 & 0.353 \scriptsize$\pm$ 0.25 \\
 AutoMetrics (Ours) & \textbf{0.316} \scriptsize$\pm$ 0.02 & \textbf{0.342} \scriptsize$\pm$ 0.01 & \textbf{0.382} \scriptsize$\pm$ 0.05 & \textbf{0.145} \scriptsize$\pm$ 0.00 & \textbf{0.365} \scriptsize$\pm$ 0.08 \\
\bottomrule
\end{tabular}
}
\caption{Criterion Validity results showing Kendall's Tau with 95\% confidence intervals over 5 independent runs.  AutoMetrics outperforms the baselines on all five tasks with Qwen3-32B and is within 95\% confidence of the best for 4/5 tasks with GPT-4o-mini.  On EvalGen, AutoMetrics improves performance by 33.4\% over the closest baseline (LLM Judge).}
\label{tab:main_results}
\end{table*}

\paragraph{AutoMetrics correlates better than all baselines across all five tasks.} We find that AutoMetrics outperforms all other existing baselines on all five tasks.  While the best performing baseline is both inconsistent on dataset (LLM Judge on \begin{wrapfigure}{r}{0.45\textwidth}
    \includegraphics[width=\linewidth]{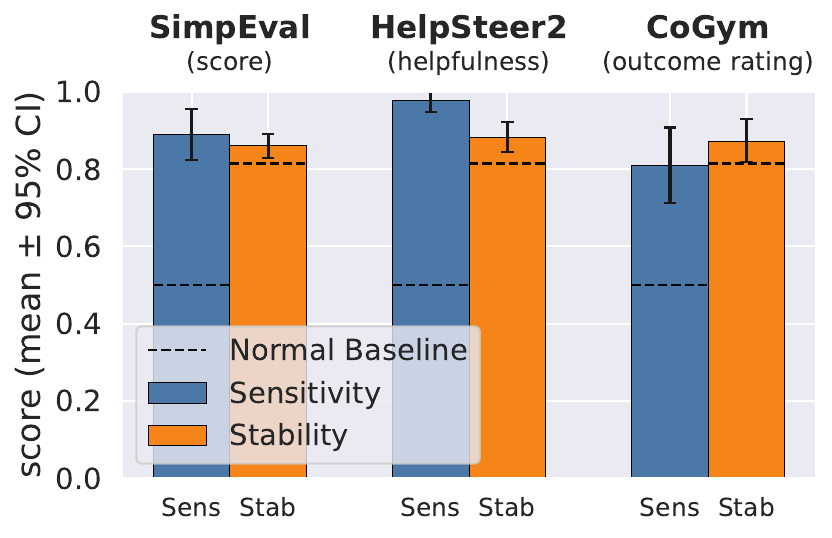}
    \caption{Sensitivity/Stability of AutoMetrics for SimpEval, HelpSteer2, and CoGym.  AutoMetrics are sensitive to negative perturbations and stable on neutral perturbations.}
    \label{fig:robustness}
    \vspace{-24pt}
\end{wrapfigure}SimpEval, HelpSteer, EvalGen; DnA Eval on RealHumanEval and CoGym) and on the underlying model used (Existing Metrics outperform GPT-4o-mini but not Qwen3-32B). In contrast, AutoMetrics is consistently the best option regardless of dataset or underlying model.  On all datasets besides HelpSteer and CoGym, the AutoMetrics performance exceeds all baselines by greater than the 95\% confidence interval. In general, AutoMetrics is the best choice for higher correlation with human ratings.

\subsection{Construct Validity (Robustness)}
\label{subsec:robustness}

To measure construct validity, we take inspiration from convergent-discriminant validity and show that AutoMetrics are strong predictors when output quality degrades and that they are stable under unimportant perturbations.  To do so we introduced \textbf{Sensitivity} and \textbf{Stability} (\S\ref{subsec:evaluation-principles}).  Sensitivity measures the rate of detection of negative perturbations and Stability measures the magnitude of score preservation under meaningless changes.  We report Sensitivity and Stability for all metrics on 30 trials in Figure~\ref{fig:robustness}.  We compare against a normal distribution baseline.

\paragraph{AutoMetrics are sensitive and stable.}  AutoMetrics are sensitive to degradation in output quality in 81.0-97.8\% of cases, depending on the dataset, which is significantly greater than the 50\% baseline. AutoMetrics can be a strong tool for identifying degradations in output quality.  Similarly, AutoMetrics also always outperforms the baseline for stability by greater than 95\% confidence intervals.  Under insignificant modifications to evaluated outputs, AutoMetrics are consistently stable. 

\subsection{Design Decisions (Hyperparameter Sweeps)}
\label{subsec:ablations}

Our sweeps/ablations test three parts of the AutoMetrics method: the MetricBank, the retrieval step, and the regression step.  We report Kendall's $\tau$ rank correlation across our six main tasks with 95\% confidence intervals over five runs in Table~\ref{tab:ablations_kendall}.  All sweeps and ablations are instead done on the dev set for all datasets.  We never make design decisions based on runs of our test sets.

\begin{table*}[h]
    \centering
    \resizebox{\textwidth}{!}{%
    \begin{tabular}{lcc|ccc}
        \toprule
        \rowcolor[gray]{0.9}
         & \multicolumn{2}{c|}{\textbf{In-Distribution}} & \multicolumn{3}{c}{\textbf{Out-of-Distribution}} \\
        \rowcolor[gray]{0.9}
        Method & SimpEval & HelpSteer2 & EvalGen & RealHumanEval & CoGym \\
        \midrule
    
        \rowcolor[gray]{0.85} \multicolumn{6}{l}{\textbf{MetricBank Ablations (k=30; n=5)}} \\
        Existing Metrics Only          & 0.238 {\scriptsize $\pm$ 0.04} & \underline{0.376 {\scriptsize $\pm$ 0.00}} & 0.389 {\scriptsize $\pm$ 0.00} & \textbf{0.155 {\scriptsize $\pm$ 0.00}} & 0.258 {\scriptsize $\pm$ 0.00} \\
        Generated Metrics Only         & \textbf{0.276 {\scriptsize $\pm$ 0.03}} & 0.308 {\scriptsize $\pm$ 0.01} & \textbf{0.503 {\scriptsize $\pm$ 0.03}} & 0.132 {\scriptsize $\pm$ 0.00} & \textbf{0.433 {\scriptsize $\pm$ 0.04}} \\
        Full MetricBank                & \underline{0.275 {\scriptsize $\pm$ 0.02}} & \textbf{0.387 {\scriptsize $\pm$ 0.00}} & \underline{0.474 {\scriptsize $\pm$ 0.03}} & \underline{0.152 {\scriptsize $\pm$ 0.01}} & \underline{0.329 {\scriptsize $\pm$ 0.02}} \\
        \midrule
        \rowcolor[gray]{0.85} \multicolumn{6}{l}{\textbf{Retrieval Ablations (n=5)}} \\
        Retrieve k=5                   & 0.257 {\scriptsize $\pm$ 0.03} & 0.336 {\scriptsize $\pm$ 0.03} & 0.414 {\scriptsize $\pm$ 0.12} & 0.124 {\scriptsize $\pm$ 0.02} & \textbf{0.385 {\scriptsize $\pm$ 0.04}} \\
        Retrieve k=10                  & 0.245 {\scriptsize $\pm$ 0.02} & 0.352 {\scriptsize $\pm$ 0.01} & 0.469 {\scriptsize $\pm$ 0.06} & 0.128 {\scriptsize $\pm$ 0.01} & \underline{0.371 {\scriptsize $\pm$ 0.02}} \\
        No Metric Cards (k=20)         & \underline{0.281 {\scriptsize $\pm$ 0.04}} & 0.328 {\scriptsize $\pm$ 0.02} & 0.427 {\scriptsize $\pm$ 0.09} & 0.134 {\scriptsize $\pm$ 0.01} & 0.292 {\scriptsize $\pm$ 0.06} \\
        Retrieve k=20                  & \textbf{0.286 {\scriptsize $\pm$ 0.02}} & \underline{0.378 {\scriptsize $\pm$ 0.01}} & \textbf{0.522 {\scriptsize $\pm$ 0.02}} & \underline{0.141 {\scriptsize $\pm$ 0.01}} & 0.302 {\scriptsize $\pm$ 0.06} \\
        Retrieve k=30                  & 0.275 {\scriptsize $\pm$ 0.02} & \textbf{0.387 {\scriptsize $\pm$ 0.00}} & \underline{0.474 {\scriptsize $\pm$ 0.03}} & \textbf{0.152 {\scriptsize $\pm$ 0.01}} & 0.329 {\scriptsize $\pm$ 0.02} \\
        \midrule
        \rowcolor[gray]{0.85} \multicolumn{6}{l}{\textbf{Regression Ablations (k=30)}} \\
        No Regression (n=1)            & 0.232 {\scriptsize $\pm$ 0.08} & \textbf{0.393 {\scriptsize $\pm$ 0.00}} & 0.353 {\scriptsize $\pm$ 0.23} & 0.145 {\scriptsize $\pm$ 0.00} & \underline{0.356 {\scriptsize $\pm$ 0.00}} \\
        Regress n=3                    & 0.255 {\scriptsize $\pm$ 0.02} & \underline{0.389 {\scriptsize $\pm$ 0.02}} & \textbf{0.503 {\scriptsize $\pm$ 0.10}} & \underline{0.152 {\scriptsize $\pm$ 0.01}} & 0.302 {\scriptsize $\pm$ 0.04} \\
        Regress n=5                    & \underline{0.275 {\scriptsize $\pm$ 0.02}} & 0.387 {\scriptsize $\pm$ 0.00} & 0.474 {\scriptsize $\pm$ 0.03} & \underline{0.152 {\scriptsize $\pm$ 0.01}} & 0.329 {\scriptsize $\pm$ 0.02} \\
        Regress n=10                   & \textbf{0.309 {\scriptsize $\pm$ 0.01}} & 0.358 {\scriptsize $\pm$ 0.01} & 0.461 {\scriptsize $\pm$ 0.05} & 0.147 {\scriptsize $\pm$ 0.01} & 0.297 {\scriptsize $\pm$ 0.05} \\
        Regress n=20                   & 0.268 {\scriptsize $\pm$ 0.03} & 0.350 {\scriptsize $\pm$ 0.01} & \underline{0.498 {\scriptsize $\pm$ 0.04}} & \textbf{0.153 {\scriptsize $\pm$ 0.01}} & \textbf{0.361 {\scriptsize $\pm$ 0.02}} \\

        \bottomrule
    \end{tabular}
    }
    \caption{Kendall correlation with 95\% confidence intervals on in-distribution and out-of-distribution datasets over five runs with Qwen3 32B (Reasoning).  The Full MetricBank and Metric Cards prove useful, and the best settings for retrieval and regression are k=30 and n=5 respectively.}
    \label{tab:ablations_kendall}
\end{table*}

\paragraph{Both Generated and Existing Metrics Help.} In all of our tasks, the Full MetricBank was either the best or second-best performing setting for the ablations.  When it was second best, it was typically within 95\% confidence intervals.  The primary exception is CoGym, where ``Full MetricBank" fell 0.104 below ``Generated Metrics Only" and, to a lesser extent, EvalGen, where ``Full MetricBank" was short by 0.029.  CoGym and EvalGen are also our smallest training sets (37 and 57 training samples respectively). We hypothesize this is because on out-of-distribution tasks, existing metrics tend to be noisy predictors which can spuriously correlate during the regression.  Generated metrics tend to be less noisy predictors.  Larger training sets provide a more effective filter for identifying useful metrics.  We further explore this hypothesis in our data scaling experiment (\S\ref{subsec:data_efficiency}).

\paragraph{Metric Cards Help Retrieval and Larger k Is Better.}  Across all five tasks, retrieval with Metric Cards (k=20) is better than retrieval without metric cards (using a single sentence description of the metric). Furthermore, we see roughly linear growth of correlation with higher $k$ metrics retrieved to run on the train set.  The single exception to this trend is CoGym, which can be attributed to the noisiness of the small dataset and the generated metrics being less noisy predictors.  The top 5 retrieved metrics are often generated ones, reducing the risk of recommending spuriously correlated existing metric on the small dataset.  We ran all retrieval experiments by regressing with $n=5$, so it is worth noting that future improvements to the retrieval algorithm (possibly including historical usage data) mean that it is feasible for $k=5$ numbers to match our $k=30$ results, so long as the proper metrics are recommended.

\paragraph{Number to regress to varies from dataset to dataset, but five is a good average case.} The best case for regression only repeats once (with n=20), suggesting that the number of metrics needed is highly dependent on the complexity of the evaluation task and domain.  Since there is no clear winner, we select n=5 as a default because it is the second best in two of five tasks, and it is the cheapest option that still maintains lower variance from run to run.  A higher N means producing more expensive metrics to run downstream, so n=5 is a useful compromise of cost and performance.

\subsection{How much data do you need to use AutoMetrics?}
\label{subsec:data_efficiency}

\begin{figure*}[h]
    \centering
    \includegraphics[width=1.0\textwidth]{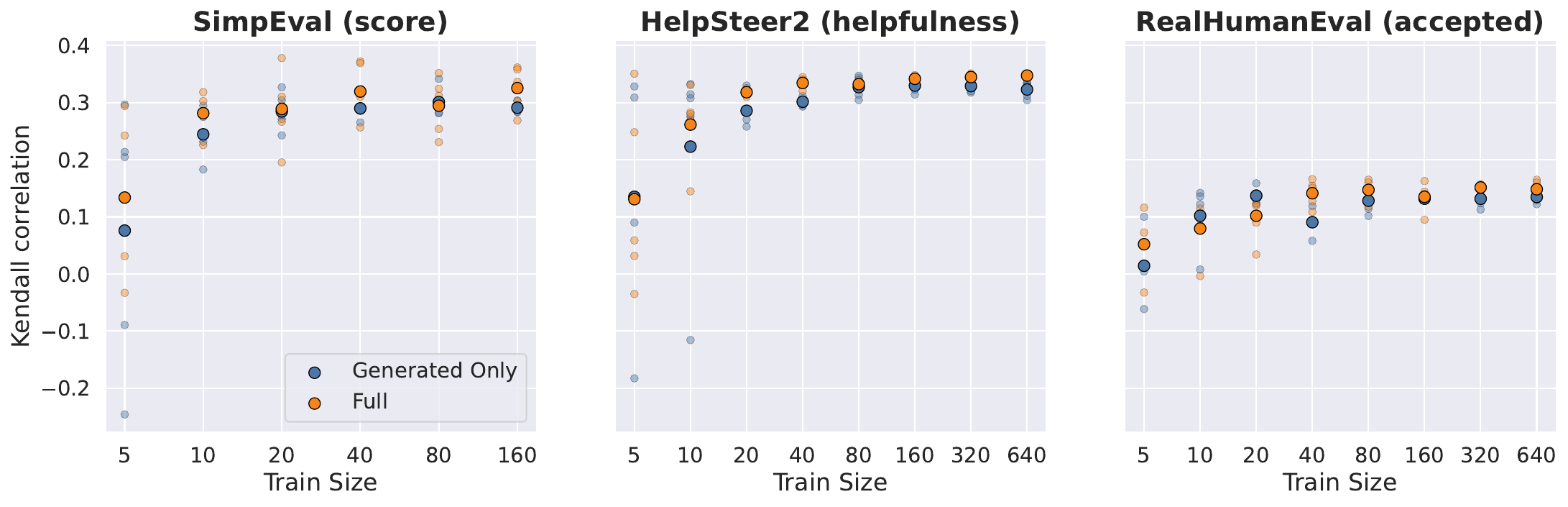}
    \caption{All correlations plotted for various training set sizes with ``Generated Only" and ``Full" Metric Banks.  Individual trials are translucent while average performance at a scale is solid.}
    \label{fig:data_efficiency}
\end{figure*}

To test how much data is needed to use AutoMetrics, we test on three distinct datasets large enough to be useful in this experiment.  We take a relatively simple \textit{In-Distribution} dataset, SimpEval, a more challenging \textit{In-Distribution} dataset, HelpSteer2, and an \textit{Out-of-Distribution} dataset RealHumanEval.  We vary the train set size from N=5, 10, 20, 40, 80, 160, and (for RealHumanEval and Helpsteer2) 320 and 640.  We run these settings for both the ``Generated Only" Metric Bank and ``Full" Metric Bank (with existing metrics). We plot the correlation on the full test set in Figure~\ref{fig:data_efficiency}.

\paragraph{About 80 samples saturates performance.}  Across all three datasets and both settings, performance levels off after about 80 samples.  It is possible with more sophisticated metric generation/learning methods more data could continue to help, however with the current architecture between 80-100 examples is all you need.  Below 80 examples most of the lower performance is due to the high variance of fitting a regression to a small training set.

\paragraph{On out-of-distribution datasets ``Generated Only" can outperform ``Full" with low-resources.}  Looking to the RealHumanEval plot we see at training size 10 and 20 the ``Generated Only" metrics outperform the ``Full" Bank.  Recall back to the ablations (\S\ref{subsec:ablations}) where we observed on the small, out-of-distribution datasets, CoGym and EvalGen, that ``Generated Only" outperformed the ``Full" MetricBank.  Since most tasks will be out of distribution by nature, we default to using ``Generated Only" when the user provides less than 80 training samples.  Beyond 80, both ``Generated Only" and ``Full" level off, however ``Full" asymptotes higher than ``Generated Only" on all datasets.  We argue this is a product of the high-p, low-n problem in regression where having too many weak predictors and not enough datapoints can lead to spurious correlations.  By limiting to generated metrics for low-n settings we enforce the use of stronger predictor signals.
\section{Case Study: AutoMetrics for Optimizing An Agentic Task}
\label{sec:case_study}

A natural extension to using AutoMetrics is to take the limited data one has available in order to learn a useful set of metrics that can then be used for optimizing a system.  In this way AutoMetrics would operate similar to the purpose of a Reward Model or a Verifiable Reward.  In order to test if AutoMetrics can be useful in this setting we optimize an airline assistance agent for $\tau$-bench \citep{yao2024tau}, a testbed for tool-use agents to interact with simulated users to accomplish tasks.  We split the 50 $\tau$-airline tasks into 25 for training and 25 for evaluation.

\begin{figure}[t]
  \centering
  \begin{minipage}[t]{0.49\columnwidth}
    \centering
    \includegraphics[width=\linewidth]{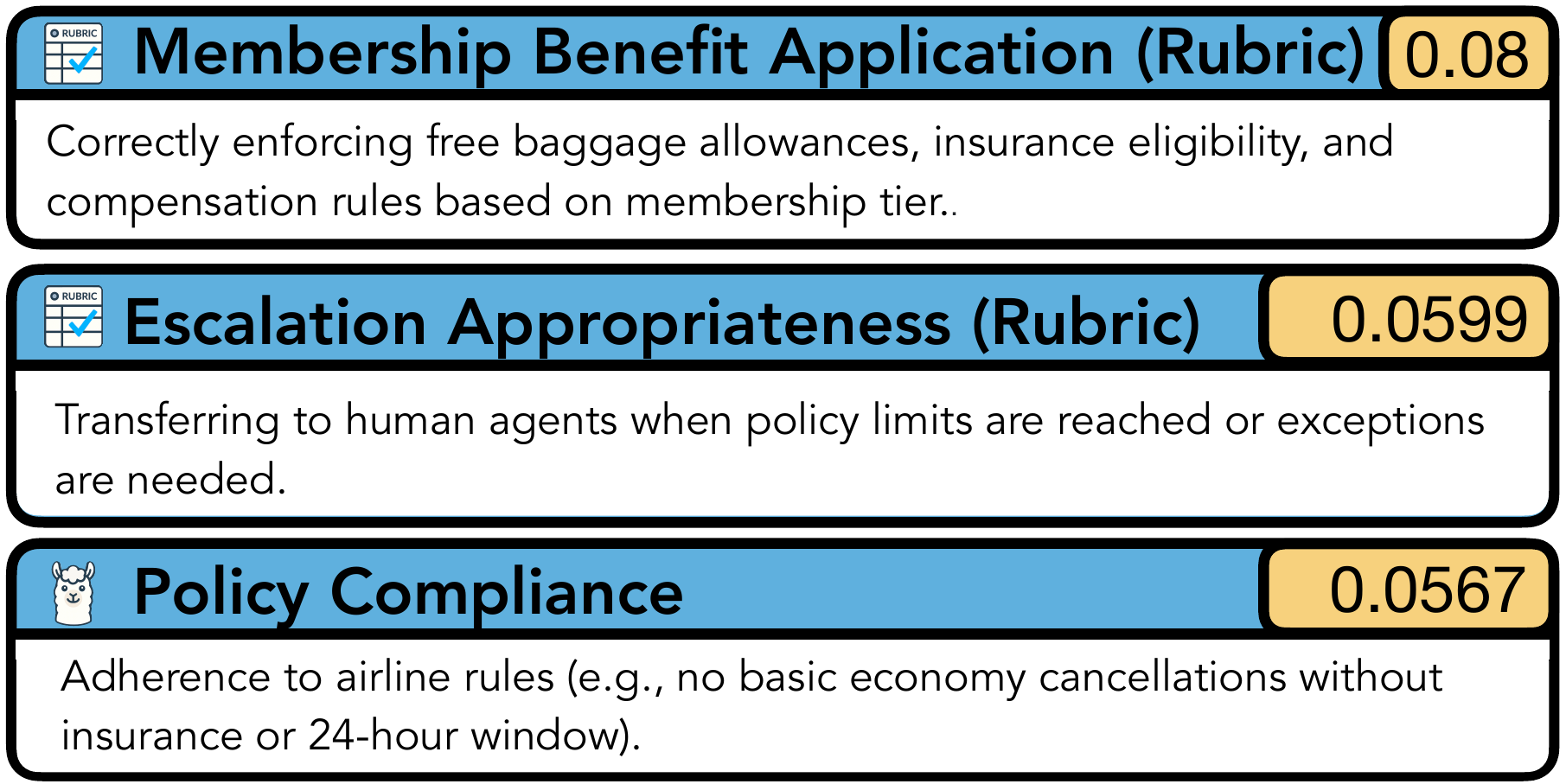}
    \caption{AutoMetrics produces three metrics for $\tau$-Bench. Regression coefficients in yellow.}
    \label{fig:tau-metrics}
  \end{minipage}
  \hfill
  \begin{minipage}[t]{0.49\columnwidth}
    \centering
    \includegraphics[width=\linewidth]{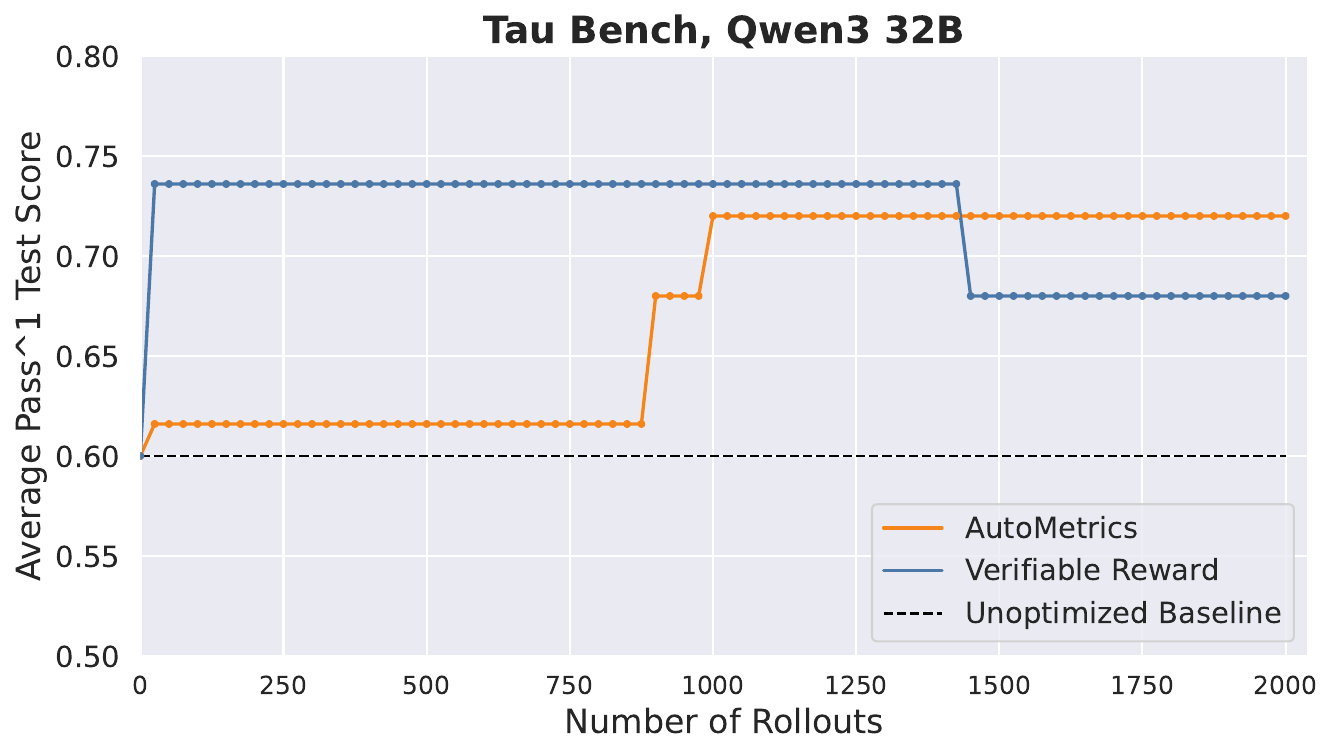}
    \caption{$\tau$-Bench performance over GEPA optimization steps when using AutoMetrics.}
    \label{fig:optimization}
  \end{minipage}
\end{figure}

\paragraph{Simulating a verifiable reward.}  To run AutoMetrics we rollout the 25 training examples 8 times each with temperatures [0.0, 0.01, 0.02, 0.03, 0.05, 0.1, 0.15, 0.2].  Then we obtain the true reward signal for each of these rollouts.  In practice rather than a verifiable reward this could be a subjective human label.  We run AutoMetrics in "Generate Only" mode and allocate more resources to generated metrics (10$\rightarrow$20 llm judge metrics; 5$\rightarrow$8 rubric metrics).  Otherwise we run with default hyperparameters (k=30; n=5).  We show the generated metrics in Figure~\ref{fig:tau-metrics}.


AutoMetrics recommends three metrics for Tau-Bench evals: two rubric based metrics and one single criterion metric.  Originally our (n=5) setting recommended five metrics, however our final filtering step removed two metrics for having negative coefficients.  Since the trajectories are only derived from 25 examples it is likely that metrics will begin to learn things about the data itself.  This reflects the importance of both human oversight and our metric filtering.



\paragraph{Optimizing without a Verifiable Reward}
We implement a simple ReAct \citep{yao2023react} agent in DSPy \citep{khattab2024dspy} for performing the $\tau$-Airline task.  Our baseline agent gets 60\% accuracy on the 25 test examples averaged over five trials.  We then run a baseline optimization where we use the DSPy GEPA optimizer \citep{agrawal2025gepareflectivepromptevolution} to optimize an agent on the 25 training tasks with \textbf{Verifiable Reward}.  Next we run optimization with our \textbf{AutoMetrics} as the metric for GEPA optimization.  We show the performance on the test set after N rollouts in Figure~\ref{fig:optimization}.  We find that \textbf{AutoMetrics can match performance of a verifiable reward.}  After 2000 rollouts the GEPA optimization with verifiable reward achieves $0.680 \pm 0.11$ accuracy over 5 trials while the AutoMetrics run gets $0.720 \pm 0.06$.  AutoMetrics statistically significantly exceeds the baseline performance ($p<0.05$) of 0.6.  This demonstrates that AutoMetrics can match or exceed Verifiable Rewards as optimization signal.

\section{Discussion and Conclusion}
\label{sec:conclusion}

In this paper, we introduced \textbf{AutoMetrics}, a method for producing metrics that correlate with human judgments on subjective tasks. Requiring only $\sim$80 human-labeled examples, AutoMetrics achieve high criterion validity (\S\ref{subsec:correlation}) and construct validity. (\S\ref{subsec:robustness}).  AutoMetrics improve upon existing baselines by up to 33.4\% in Kendall correlation with human ratings. In a case study on Tau-Bench, AutoMetrics matched or exceeded gains obtained from optimizing on a verifiable reward (\S\ref{sec:case_study}).

We draw two key lessons for practitioners. First, \textbf{data diversity is critical}: while only $\sim$80 feedback points suffice for moderate correlation (\S\ref{subsec:data_efficiency}), scaling up synthetic data from limited sources can produce metrics that reflect dataset artifacts rather than system quality (\S\ref{sec:case_study}). Second, \textbf{human oversight remains essential}: domain experts can help remove spuriously correlated metrics which the automatic filtering process misses. When using metrics for optimization, practitioners should monitor metric feedback and improvement with observability tools \citep{langwatch}.

Overall, \textbf{AutoMetrics} provides a practical first step for exploring data and guiding optimization when collecting preliminary human evaluation in new domains. The metrics it produces are interpretable, actionable, and informative for system improvement. We release AutoMetrics publicly and invite community contributions of new metrics and methods to strengthen the framework.

\section*{Reproducibility Statement}
\label{sec:reproducibility}

AutoMetrics is intended to be an open source library and framework.  As such we take great effort to make the running and evaluation of AutoMetrics user-friendly.  We have attached an anonymized repository for AutoMetrics with this submission.  In addition to the core algorithm, the repository also contains the python scripts to reproduce all experimental results in this paper.  All of our design decisions, hyperparameters, and ablations are rigorously documented throughout the paper across Section~\ref{subsec:ablations} and Appendix~\ref{appendix:autometrics_design_ablations}.  We provide system-specs needed to run the metrics in Table~\ref{tab:ci_resource_usage}.  We also share the exact prompts and DSPy signatures used in calling LLMs in Appendix~\ref{appendix:prompts}.  For all main experimental results (e.g. Table~\ref{tab:main_results} and Table~\ref{tab:ablations_kendall}) results are reported over five independent random seeds to ensure findings are robust and statistically significant.

\section*{Limitations}
\label{sec:limitations}

As a part of the AutoMetrics framework we construct and optimize metrics with particular LLMs.  Because the metric generation process involves optimizing to a particular model we have found that producing metrics with one model and running them with another reduces performance.  This suggests that when better models are released it will be important to reoptimize automatic metrics using AutoMetrics rather than just swap out the underlying LLM.

AutoMetrics may only generalize as far as the provided data enables it.  Collecting real, diverse human data is still an essential part of evaluation.  The more representative and generalizable the input data is, the better and more general the AutoMetrics will be.  Users should collect data that is representative of the opinions and population that they want their evaluation to cover.


AutoMetrics depends on running a regression for many predictors on a limited number of data points.  Although we took this into account with the design of our Regression step, it is still possible to run into a high-P low-N regression problem that risks spurious correlations.  To counteract accidental misuse of AutoMetrics leading to poor evaluation, we add warnings to the metric reports when the significance of the correlation with human judgments of the recommended metric is low ($p > 0.05$).

Finally, as a part of this work we do not conduct a formal user study to demonstrate the adoption of AutoMetrics among practitioners.  We have collected positive feedback on the metrics through informal tests with AI developers. We hope that by releasing and open sourcing this library, we will have the opportunity to work with the community to test and improve AutoMetrics.
\section*{Acknowledgements}
This work has been supported in part through the Stanford HAI Corporate Affiliate Program, with membership funding provided by American Express.  This work was also funded through a grant from the Sloan Foundation.  The authors would like to thank Omar Khattab, William Held, Saurabh Shah, Aryaman Arora, Ken Liu, David Anugraha, Vishakh Padmakumar, Hao Zhu, Lakshya Agrawal, Yu Fei, Seungone Kim, and Jonathan Hilgart for their insightful comments and thoughts at various stages of the project.  We would also like to thank SALT Lab and the Stanford NLP Group for help with review and revision of the manuscript.  Finally we would like to thank Yijia Shao and Alex Spangher for testing and offering feedback on the AutoMetrics system.



\bibliography{iclr2026_conference}
\bibliographystyle{iclr2026_conference}

\appendix

\section{LLM Usage Acknowledgment}
\label{appendix:llm_acknowkedgment}

LLMs were used to rephrase and edit writing in the paper, after an entirely human-written first draft.  LLMs were also used as coding assistants in writing code for this project.  All code and writing edits produced by LLMs were rigorously verified by the first-author.
\section{Introducing MetricBank}
\paragraph{MetricBank}
\label{sec:metric_bank}

As our first significant contribution, we curate MetricBank, a standardized collection of 48 commonly used metrics in NLP literature.  We source the metrics from \citet{schmidtova-etal-2024-automatic-metrics}, which examined all papers from the \textit{International Conference on Natural Language Generation} (INLG) 2023 and all papers in the \textit{Generation} track presented at ACL 2023, totaling 110 papers.  They collected a list of all the Natural Language Generation (NLG) metrics used in those works, which totaled 283 different automatic metrics grouped into 34 metric families.  We sorted by the most popular and implemented the top metrics from the top 16 families (28 metrics).  Then, for completeness, we also implemented any remaining NLG metrics in NLTK \cite{bird-loper-2004-nltk}, PyTorch \cite{10.5555/3454287.3455008}, Huggingface Lighteval \cite{lighteval}, and Metametrics \cite{winata2025metametrics} for an additional 12 metrics.  Finally, we source a few additional metrics from recent papers not covered in the 2024 survey.  We provide individual justifications for these 8 metrics in Appendix~\ref{appendix:additional_metrics}.

We provide interesting stats about our metrics in Table~\ref{tab:metrics}.  In particular, we collect 29 reference-based metrics, such as BLEU \cite{papineni-etal-2002-bleu}, which require a gold reference output, and 19 reference-free metrics, such as FKGL \cite{flesch_marks_1943}, which measure quality of text without comparison to a reference.  Our metrics span 12 distinct domains.  We implement each metric with a simple interface of a \texttt{calculate} method that takes in the generated text and produces a floating-point score and optionally text feedback.

\paragraph{Metric Cards}
\label{sec:metric_cards}
Inspired by Model Cards \cite{10.1145/3287560.3287596} and Data Cards \cite{10.1145/3531146.3533231} we design Metric Cards for simple documentation and reporting of the intended usage of metrics.  Our Metric Cards contains seven main sections. \textbf{Metric Details} contains the description of the metric as well as core details that are needed to use it, such as the range of outputs, if it's reference based, if an input is required, etc. \textbf{Intended Use} describes the domain/tasks where the metric should be used as well as recommendations for when and when not to use the metric.  \textbf{Metric Implementation} links to reference implementations and provides guidance on practical matters about the metric such as it's efficiency and scalability.  \textbf{Known Limitations} explains biases, misuse, and known failure cases of the metric. \textbf{Related Metrics} links to similar metrics to help when browsing for the right metric for your task.  \textbf{Further Reading} points to papers, blogs, and tutorials covering the metric.  Finally, \textbf{Metric Card Authors} makes it clear who wrote the metric card and if they used an AI assistance.  We provide a complete example of a metric card for the common BLEU metric \cite{papineni-etal-2002-bleu} in Appendix~\ref{appendix:example_metric_card}.  We also provide a prompt for using LLMs to write a first pass of a metric card in Appendix~\ref{appendix:prompts}.

\begin{table*}[h!]
  \centering
  \small
  \setlength{\tabcolsep}{6pt}
  \renewcommand{\arraystretch}{1.1}
  \resizebox{\textwidth}{!}{%
    \begin{tabular}{l c c l c l}
      \toprule
      \rowcolor{gray!30}
      \textbf{Metric (Citation)} 
        & \textbf{Domain} 
        & \textbf{GPU} 
        & \textbf{Type} 
        & \textbf{Sup.} 
        & \textbf{Default LLM} \\
      \midrule
      \rowcolor{gray!10}
      \multicolumn{6}{l}{\textit{\textbf{Reference-Based Metrics}: rely on a gold reference for comparison.}} \\
      \midrule
      Jaccard Distance \cite{jaccard1901etude}                   
        & \faIcon{file-alt},\,\faIcon{retweet},\,\faIcon{edit}             
        & \xmark 
        & edit-distance     
        & \xmark 
        & N.A.               \\
      Hamming Distance \cite{hamming1950error}                   
        & \faIcon{edit}                                                   
        & \xmark 
        & edit-distance     
        & \xmark 
        & N.A.               \\
      Levenshtein Distance \cite{levenshtein1965binary}          
        & \faIcon{globe},\,\faIcon{file-alt},\,\faIcon{retweet},\,\faIcon{edit} 
        & \xmark 
        & edit-distance     
        & \xmark 
        & N.A.               \\
      Levenshtein Ratio \cite{levenshtein1965binary}             
        & \faIcon{globe},\,\faIcon{file-alt},\,\faIcon{retweet},\,\faIcon{edit} 
        & \xmark 
        & edit-distance     
        & \xmark 
        & N.A.               \\
      Jaro Similarity \cite{3ebd50ca-85b6-3914-bf38-759fcad3ed72}  
        & \faIcon{edit}                                                   
        & \xmark 
        & edit-distance     
        & \xmark 
        & N.A.               \\
      Jaro–Winkler \cite{winkler1990string}                      
        & \faIcon{edit}                                                   
        & \xmark 
        & edit-distance     
        & \xmark 
        & N.A.               \\
      BLEU \cite{papineni-etal-2002-bleu}                        
        & \faIcon{globe}                                                  
        & \xmark 
        & n-gram overlap    
        & \xmark 
        & N.A.               \\
      NIST \cite{10.5555/1289189.1289273}                         
        & \faIcon{globe},\,\faIcon{file-alt}                              
        & \xmark 
        & n-gram overlap    
        & \xmark 
        & N.A.               \\
      ROUGE \cite{lin-2004-rouge}                                
        & \faIcon{file-alt},\,\faIcon{globe},\,\faIcon{retweet},\,\faIcon{file-invoice} 
        & \xmark 
        & n-gram overlap    
        & \xmark 
        & N.A.               \\
      METEOR \cite{banerjee-lavie-2005-meteor}                   
        & \faIcon{globe},\,\faIcon{file-alt},\,\faIcon{retweet},\,\faIcon{image} 
        & \xmark 
        & n-gram overlap    
        & \xmark 
        & N.A.               \\
      TER \cite{snover-etal-2006-study}                          
        & \faIcon{globe}                                                  
        & \xmark 
        & edit-distance     
        & \xmark 
        & N.A.               \\
      iBLEU \cite{sun-zhou-2012-joint}                           
        & \faIcon{retweet},\,\faIcon{book-reader},\,\faIcon{comments}     
        & \xmark 
        & n-gram overlap    
        & \xmark 
        & N.A.               \\
      CHRF++ \cite{popovic-2015-chrf}                            
        & \faIcon{globe},\,\faIcon{file-alt},\,\faIcon{retweet},\,\faIcon{file-invoice} 
        & \xmark 
        & n-gram overlap    
        & \xmark 
        & N.A.               \\
      CIDEr \cite{7299087}                                       
        & \faIcon{image}                                                  
        & \xmark 
        & n-gram overlap    
        & \xmark 
        & N.A.               \\
      GLEU \cite{wu2016googlesneuralmachinetranslation}          
        & \faIcon{globe}                                                  
        & \xmark 
        & n-gram overlap    
        & \xmark 
        & N.A.               \\
      SARI \cite{xu-etal-2016-optimizing}                        
        & \faIcon{file-alt},\,\faIcon{book-reader}                        
        & \xmark 
        & n-gram overlap    
        & \xmark 
        & N.A.               \\
      CharCut \cite{lardilleux-lepage-2017-charcut}              
        & \faIcon{globe},\,\faIcon{edit}                                  
        & \xmark 
        & edit-distance     
        & \xmark 
        & N.A.               \\
      MoverScore \cite{zhao-etal-2019-moverscore}                
        & \faIcon{globe},\,\faIcon{file-alt},\,\faIcon{image},\,\faIcon{file-invoice} 
        & \cmark 
        & embedding sim     
        & \xmark 
        & BERT               \\
      PseudoPARENT \cite{dhingra-etal-2019-handling}             
        & \faIcon{file-invoice},\,\faIcon{file-alt},\,\faIcon{retweet}    
        & \xmark 
        & n-gram overlap    
        & \xmark 
        & N.A.               \\
      BERTScore \cite{Zhang2020BERTScore}                        
        & \faIcon{globe},\,\faIcon{file-alt},\,\faIcon{retweet},\,\faIcon{image} 
        & \cmark 
        & embedding sim     
        & \xmark 
        & RoBERTa-Large      \\
      BLEURT \cite{sellam-etal-2020-bleurt}                      
        & \faIcon{globe},\,\faIcon{file-alt},\,\faIcon{retweet},\,\faIcon{file-invoice} 
        & \cmark 
        & LM regression     
        & \cmark 
        & BERT/RemBERT       \\
      BARTScore \cite{10.5555/3540261.3542349}                    
        & \faIcon{globe},\,\faIcon{file-alt},\,\faIcon{retweet},\,\faIcon{file-invoice},\,\faIcon{comments} 
        & \cmark 
        & LM regression     
        & \xmark 
        & BART               \\
      InfoLM \cite{Colombo2021InfoLMAN}                          
        & \faIcon{file-alt},\,\faIcon{file-invoice}                      
        & \cmark 
        & divergence-based  
        & \xmark 
        & BERT               \\
      MAUVE \cite{pillutla-etal:mauve:neurips2021}               
        & \faIcon{comments},\,\faIcon{feather-alt}                       
        & \cmark 
        & divergence-based  
        & \xmark 
        & GPT-2              \\
      ParaScore \cite{shen-etal-2022-evaluation}                 
        & \faIcon{retweet}                                               
        & \cmark 
        & embedding sim     
        & \xmark 
        & RoBERTa-large      \\
      UniEvalDialogue \cite{zhong-etal-2022-towards}             
        & \faIcon{comments}                                              
        & \cmark 
        & LM regression     
        & \cmark 
        & T5                 \\
      UniEvalSum \cite{zhong-etal-2022-towards}                  
        & \faIcon{file-alt},\,\faIcon{file-invoice}                      
        & \cmark 
        & LM regression     
        & \cmark 
        & T5                 \\
      UpdateROUGE \cite{iv-etal-2022-fruit}                      
        & \faIcon{file-alt}                                              
        & \xmark 
        & n-gram overlap    
        & \xmark 
        & N.A.               \\
      LENS \cite{maddela-etal-2023-lens}                        
        & \faIcon{file-alt},\,\faIcon{book-reader}                      
        & \cmark 
        & LM regression     
        & \cmark 
        & T5                 \\
      \midrule
      \rowcolor{gray!10}
      \multicolumn{6}{l}{\textit{\textbf{Reference-Free Metrics}: do not require a gold reference.}} \\
      \midrule
      FKGL \cite{kincaid1975derivation}                          
        & \faIcon{file-alt},\,\faIcon{book-reader}                      
        & \xmark 
        & rule-based        
        & \xmark 
        & N.A.               \\
      Perplexity \cite{10.1121/1.2016299}                        
        & \faIcon{comments},\,\faIcon{feather-alt},\,\faIcon{code}      
        & \cmark 
        & fluency           
        & \xmark 
        & GPT-2 Large              \\
      DistinctNGrams \cite{li-etal-2016-diversity}               
        & \faIcon{comments},\,\faIcon{feather-alt}                      
        & \xmark 
        & diversity ratio   
        & \xmark 
        & N.A.               \\
      SelfBLEU \cite{zhu2018texygen}                            
        & \faIcon{comments},\,\faIcon{feather-alt},\,\faIcon{retweet}   
        & \xmark 
        & diversity ratio   
        & \xmark 
        & N.A.               \\
      YiSi-2 \cite{lo-2019-yisi}                                 
        & \faIcon{globe}                                                 
        & \cmark 
        & embedding sim     
        & \xmark 
        & mBERT              \\
      SummaQA \cite{scialom-etal-2019-answers}                   
        & \faIcon{file-alt}                                             
        & \cmark 
        & LM regression     
        & \cmark 
        & BERT               \\
      FactCC \cite{kryscinski-etal-2020-evaluating}              
        & \faIcon{file-alt}                                             
        & \cmark 
        & LM regression     
        & \cmark 
        & BERT               \\
      Toxicity \cite{vidgen-etal-2021-learning}                  
        & \faIcon{comments},\,\faIcon{feather-alt},\,\faIcon{shield-alt} 
        & \cmark 
        & classification    
        & \cmark 
        & RoBERTa            \\
      ParaScoreFree \cite{shen-etal-2022-evaluation}             
        & \faIcon{retweet},\,\faIcon{file-invoice}                     
        & \cmark 
        & embedding sim     
        & \xmark 
        & RoBERTa-large      \\
      Sentiment \cite{camacho-collados-etal-2022-tweetnlp}       
        & \faIcon{comments},\,\faIcon{feather-alt}                      
        & \cmark 
        & classification    
        & \cmark 
        & RoBERTa            \\
      UniEvalFact \cite{zhong-etal-2022-towards}                 
        & \faIcon{file-alt},\,\faIcon{file-invoice}                     
        & \cmark 
        & LM regression     
        & \cmark 
        & T5                 \\
      LENS\_SALSA \cite{heineman-etal-2023-dancing}              
        & \faIcon{file-alt},\,\faIcon{book-reader}                     
        & \cmark 
        & LM regression     
        & \cmark 
        & T5                 \\
      FastTextEducationalValue \cite{ktsui2024cpueduvalue}      
        & \faIcon{file-alt},\,\faIcon{book-reader}                     
        & \xmark 
        & classification    
        & \cmark 
        & FastText           \\
      FastTextNSFW \cite{soldaini-etal-2024-dolma}               
        & \faIcon{comments},\,\faIcon{shield-alt}                      
        & \xmark 
        & classification    
        & \cmark 
        & FastText           \\
      FastTextToxicity \cite{soldaini-etal-2024-dolma}           
        & \faIcon{comments},\,\faIcon{shield-alt}                      
        & \xmark 
        & classification    
        & \cmark 
        & FastText           \\
      GRMRewardModel \cite{yang2024regularizing}                
        & \faIcon{comments},\,\faIcon{shield-alt}                      
        & \cmark 
        & LM regression     
        & \cmark 
        & Llama-3.2-3B       \\
      INFORM Reward Model 70B \cite{INF-ORM-Llama3.1-70B}        
        & \faIcon{comments}                                             
        & \cmark 
        & LM regression     
        & \cmark 
        & Llama-3.1-70B      \\
      LDL Reward Model 27B \cite{chen2025ldl}                   
        & \faIcon{comments},\,\faIcon{code}                            
        & \cmark 
        & LM regression     
        & \cmark 
        & Gemma 2-27B        \\
      MathProcessRewardModel \cite{zhang2025lessonsdevelopingprocessreward} 
        & \faIcon{calculator},\,\faIcon{file-alt}                      
        & \cmark 
        & classification    
        & \cmark 
        & Qwen2.5 7B         \\
      \bottomrule
    \end{tabular}%
  }
  \caption{Comparison of generative evaluation metrics.  
    \textbf{Icons:}
    \faIcon{globe} Machine Translation,
    \faIcon{file-alt} Summarization,
    \faIcon{retweet} Paraphrasing,
    \faIcon{comments} Dialogue/Chat,
    \faIcon{feather-alt} Storytelling/Creative Writing,
    \faIcon{image} Image Captioning/Multimodal,
    \faIcon{shield-alt} Safety/Moderation,
    \faIcon{file-invoice} Data-to-Text Generation,
    \faIcon{book-reader} Education/Readability,
    \faIcon{code} Code Generation,
    \faIcon{calculator} Math/Problem Solving,
    \faIcon{edit} String-Distance/Edit-Based.}
  \label{tab:metrics}
\end{table*}

\begin{table*}[h!]
  \centering
  \small
  \setlength{\tabcolsep}{6pt}
  \renewcommand{\arraystretch}{1.1}
  \resizebox{0.8\textwidth}{!}{%
    \begin{tabular}{l r r r}
      \toprule
      \rowcolor{gray!30}
      \textbf{Metric} & \textbf{GPU} & \textbf{CPU} & \textbf{Time (ms)} \\
      \midrule
      INFORMRewardModel                     & 129.62\,GB & 2.04\,GB & 1041   \\
      LDLRewardModel & 104.17\,GB & 2.06\,GB & 1921 \\
      GRMRewardModel                        &   6.02\,GB & 1.96\,GB & 61    \\
      UniEvalDialogue                       &   3.07\,GB & 3.10\,GB & 262   \\
      UniEvalSum                            &   3.07\,GB & 3.10\,GB & 211   \\
      UniEvalFact                           &   3.07\,GB & 3.09\,GB & 61    \\
      Perplexity\_gpt2-large                &   3.00\,GB & 1.47\,GB & 48    \\
      BLEURT                                &   2.15\,GB & 2.75\,GB & 43    \\
      BARTScore\_bart-large-cnn             &   1.52\,GB & 1.34\,GB & 49    \\
      SummaQA                               &   1.25\,GB & 1.51\,GB & 879   \\
      YiSi                                  & 687\,MB    & 1.39\,GB & 35    \\
      Sentiment                             & 485\,MB    & 1.36\,GB & 19    \\
      Toxicity                              & 485\,MB    & 1.36\,GB & 39    \\
      FactCC                                & 427\,MB    & 1.29\,GB & 17    \\
      ParaScoreFree                         & 346\,MB    & 1.68\,GB & 12\,428 \\
      ParaScore                             & 338\,MB    & 1.05\,GB & 4\,543 \\
      MOVERScore\_distilbert-base-uncased   & 262\,MB    & 1.50\,GB & 2\,899 \\
      BERTScore\_roberta-large              &   8\,MB    & 1.47\,GB & 1\,303 \\
      PRMRewardModel                        &   0\,MB    & 13.64\,GB & 6\,359 \\
      MAUVE\_max                            &   0\,MB    &  4.22\,GB & 3\,236 \\
      FastTextEducationalValue              &   0\,MB    &  3.73\,GB & 6     \\
      LENS                                  &   0\,MB    &  3.25\,GB & 3\,408 \\
      LENS\_SALSA                           &   0\,MB    &  2.84\,GB & 426   \\
      FastTextToxicity                      &   0\,MB    &  1.67\,GB & 11    \\
      FastTextNSFW                          &   0\,MB    &  1.67\,GB & 6     \\
      InfoLM                                &   0\,MB    &  1.12\,GB & 2\,338 \\
      METEOR                                &   0\,MB    &  1.08\,GB & 27    \\
      FKGL                                  &   0\,MB    & 894\,MB   & 6     \\
      TER                                   &   0\,MB    & 731\,MB   & 26\,064 \\
      CHRF                                  &   0\,MB    & 730\,MB   & 36    \\
      DistinctNGram                         &   0\,MB    & 730\,MB   & 19    \\
      iBLEU                                 &   0\,MB    & 730\,MB   & 18    \\
      BLEU                                  &   0\,MB    & 729\,MB   & 7     \\
      LevenshteinDistance\_min              &   0\,MB    & 729\,MB   & 0     \\
      SelfBLEU                              &   0\,MB    & 729\,MB   & 6     \\
      HammingDistance\_min                  &   0\,MB    & 729\,MB   & 0     \\
      JaroWinklerSimilarity\_max            &   0\,MB    & 729\,MB   & 0     \\
      GLEU                                  &   0\,MB    & 728\,MB   & 9     \\
      SARI                                  &   0\,MB    & 728\,MB   & 95    \\
      JaccardDistance\_min                  &   0\,MB    & 728\,MB   & 0     \\
      CharCut                               &   0\,MB    & 728\,MB   & 1\,237 \\
      UpdateROUGE                           &   0\,MB    & 728\,MB   & 96    \\
      NIST                                  &   0\,MB    & 728\,MB   & 21    \\
      LevenshteinRatio\_max                 &   0\,MB    & 727\,MB   & 0     \\
      JaroSimilarity\_max                   &   0\,MB    & 727\,MB   & 0     \\
      ROUGE                                 &   0\,MB    & 726\,MB   & 487   \\
      CIDEr\_n4\_sig6.0                     &   0\,MB    & 726\,MB   & 31    \\
      PseudoPARENT                          &   0\,MB    & 726\,MB   & 10    \\
      \bottomrule
    \end{tabular}%
  }
  \caption{Maximum CI upper-bound GPU/CPU memory and latency per metric.}
  \label{tab:ci_resource_usage}
\end{table*}

\subsection{Additional Metrics}
\label{appendix:additional_metrics}

Here we provide justifications for our additional metrics that we did not collect from the metric survey \citep{schmidtova-etal-2024-automatic-metrics}, lighteval \citep{lighteval}, torchmetrics \citep{torchmetrics}, etc.

\paragraph{Reward Models.} We choose some of the most performant reward models off the RewardBench leaderboard \citep{lambert2024rewardbenchevaluatingrewardmodels} at development time.  The three models we used are \textbf{INFORMRewardModel} (llama 3.1 70b) \citep{INF-ORM-Llama3.1-70B}, \textbf{LDLRewardModel}  (Gemma 2 27B) \citep{chen2025ldl}, and \textbf{GRMRewardModel} (Llama 3.2 3B) \citep{yang2024regularizing}.  We also add in the Qwen2.5 7B \textbf{Process Reward Model} \citep{zhang2025lessonsdevelopingprocessreward}.

\paragraph{SALSA.}  In researching text simplification metrics we found that an extension to the LENS \citep{maddela-etal-2023-lens} metric exists which is meant to better align with human judgement.  It was also a related metric to the SimpEval \citep{maddela-etal-2023-lens} paper.  Thus we chose to implement \textbf{SALSA} \citep{heineman-etal-2023-dancing} in our MetricBank as it was intended as one of the recommended ``Best" metrics for our \emph{in-distribution} SimpEval task.

\paragraph{FastText Classifiers.}  We wanted to add diversity to our MetricBank by including more classifiers for various higher-level concepts, but we didn't want to add unnecessary expenses to running the metrics.  FastText Classifiers are a nice compromise which are quick to run on CPU but also have reasonable classification accuracy.  We implement \textbf{FastTextNSFW} and \textbf{FastTextToxicity} from Dolma \citep{soldaini-etal-2024-dolma}, and we take \textbf{FastTextEducationalValue} \citep{ktsui2024cpueduvalue} which has been used for data filtering to attempt to find Text-Book quality training data.
\section{Prompts and Signatures}
\label{appendix:prompts}

\subsection{LLM-as-a-Judge Prompts}

We use the LLM-as-a-Judge Prompts from the original human annotation process for a given dataset whenever available.  We consider these as a strong baseline as these instructions were designed to be useful instructions for human annotators and ideally were the underlying instructions guiding their annotation decisions.

\begin{PromptBox}[Task: SimpEval]

\medskip
\textbf{LLM-as-a-Judge Prompt:}

\begin{lstlisting}[style=prompt,backgroundcolor=\color{PromptBG}]
## Rating Sentences

The goal is to **rate sentences** by how well they **simplify the original sentence**.

### Scoring Guidelines

| Score | When to assign it |
|-------|------------------|
| **100** | The sentence is **fully simplified**, entirely fluent, and **preserves the core meaning** of the original. |
| **75**  | The sentence is **somewhat simpler**, mostly fluent, and the meaning is **close** to the original. |
| **50**  | The sentence is simpler, **somewhat fluent**, and the meaning is **similar** to the original. |
| **25**  | The sentence is equivalently simple, still has some fluency, but **loses the meaning**. |
| **0**   | The sentence is **completely unreadable**. |

> **Most scores will lie somewhere in this range - feel free to give specific scores (e.g., 83, 67) rather than only the five anchors.**

---

### Examples

| Score | Example Simplified Sentence | Why this score? |
|-------|-----------------------------|-----------------|
| **100** | *It will then **move away from the river bed** and sink back to the bottom to digest its food.* | Reads fluently **and** keeps the original meaning ("it" gets unstuck, moves down, digests food). |
| **75** | *Due to this, **a lot of mosques don't enforce these rules** but both men and women should follow them.* | Minor fluency issue, but meaning matches the original. |
| **0** | *A gadget javascript a is and / checking wikipedia an snippet that can be enabled simply by , or css option in your wikipedia preferences.* | Sentence is **unreadable**. |
\end{lstlisting}
\end{PromptBox}

\begin{PromptBox}[Task: HelpSteer2]

\medskip
\textbf{LLM-as-a-Judge Prompt:}

\begin{lstlisting}[style=prompt,backgroundcolor=\color{PromptBG}]
**Helpfulness/Understanding:**
- 4 - The response is extremely helpful and completely aligned with the spirit of what the prompt
was asking for.
- 3 - The response is mostly helpful and mainly aligned with what the user was looking for, but
there is still some room for improvement.
- 2 - The response is partially helpful but misses the overall goal of the user's query/input in some
way. The response did not fully satisfy what the user was looking for.
- 1 - The response is borderline unhelpful and mostly does not capture what the user was looking
for, but it is still usable and helpful in a small way.
- 0 - The response is not useful or helpful at all. The response completely missed the essence of
what the user wanted.
\end{lstlisting}
\end{PromptBox}

\begin{PromptBox}[Task: EvalGenProduct]

\medskip
\textbf{LLM-as-a-Judge Prompt:}

\begin{lstlisting}[style=prompt,backgroundcolor=\color{PromptBG}]
Is this response good (1) or bad (0)?
\end{lstlisting}
\end{PromptBox}

\begin{PromptBox}[Task: RealHumanEval]

\medskip
\textbf{LLM-as-a-Judge Prompt:}

\begin{lstlisting}[style=prompt,backgroundcolor=\color{PromptBG}]
Would you accept this code edit/addition (1) or reject it (0)?
\end{lstlisting}
\end{PromptBox}

\begin{PromptBox}[Task: CoGymTravelOutcome]

\medskip
\textbf{LLM-as-a-Judge Prompt:}

\begin{lstlisting}[style=prompt,backgroundcolor=\color{PromptBG}]
Overall rating to the final outcome (i.e., travel plan, analysis result) (1-5 scale)

(1) "Extremely dissatisfied", 
(2) "Somewhat dissatisfied",
(3) "Neutral",
(4) "Somewhat satisfied",
(5) "Extremely satisfied"
\end{lstlisting}
\end{PromptBox}

\subsection{Misc Prompts}

\begin{PromptBox}[MetricCard Generation]

\medskip
\textbf{Prompt:}

\begin{lstlisting}[style=prompt,backgroundcolor=\color{PromptBG}]
You are an expert in natural language processing and technical documentation, specializing in metrics for evaluating generative models. I am building a metric bank to recommend the best metrics for various generative tasks. Each metric in this bank will have a corresponding Metric Card, which provides standardized, detailed documentation about the metric. These Metric Cards will serve as a key resource for researchers and practitioners, helping them select the right metric for their task.

## Your Task

Using the provided materials, including the original paper, reference implementations, the Metric Card Template, and the BLEU Metric Card Example, your task is to draft a comprehensive Metric Card for the given metric. The documentation must:
	1.	Follow the provided template closely, ensuring adherence to its format and required sections.
	2.	Incorporate relevant details from the original paper and reference materials, ensuring technical accuracy and completeness.
	3.	Match the style and quality of the BLEU example, which serves as an exemplar for clarity, structure, and precision.

Specific Instructions
	1.	Key Sections to Address: Ensure each required section of the template is filled out thoughtfully and thoroughly, including:
		-	Metric Description
		-	Inputs and Outputs
		-	Formal Definition
		-	Applicability and Limitations
		-	Known Limitations and Related Metrics
	2.	If Information is Unclear or Missing: Do not fabricate or make assumptions. If information is unavailable, unclear, or not included in the provided context, leave that section blank or mark it as "Needs more information."
	3.	Markdown Formatting: Output the completed Metric Card as a markdown text block rather than rendering or printing the markdown directly.  This means you must surround your answer in ```.  Also start the block with "---" as shown in the examples.  Do not end the block with "---".
	4.	Focus on Consistency: Use the provided categorical suggestions (see below) to ensure uniformity across all Metric Cards, particularly in fields like "Metric Type," "Domain," and "Tasks."
	5.	Mathematical Formatting:
		-	Use $ for inline math expressions (e.g., $r$, not $ r $).
		-	Use $$ for block math expressions and ensure a full line break before and after each block math expression. This formatting ensures proper rendering in markdown.
		-	Example of proper usage for $$:  

** Correct **
```
Where:
- $CHRP$ is the average precision of character and word n-grams:

$$
CHRP = \frac{1}{N} \sum_{n=1}^N \frac{\text{n-grams in hypothesis and reference}}{\text{total n-grams in hypothesis}}
$$

- $CHRR$ is the average recall of character and word n-grams:

$$
CHRR = \frac{1}{N} \sum_{n=1}^N \frac{\text{n-grams in hypothesis and reference}}{\text{total n-grams in reference}}
$$
```

** Incorrect **
```
Where:
- $CHRP$ is the average precision of character and word n-grams:
  $$
  CHRP = \frac{1}{N} \sum_{n=1}^N \frac{\text{n-grams in hypothesis and reference}}{\text{total n-grams in hypothesis}}
  $$
- $CHRR$ is the average recall of character and word n-grams:
  $$
  CHRR = \frac{1}{N} \sum_{n=1}^N \frac{\text{n-grams in hypothesis and reference}}{\text{total n-grams in reference}}
  $$
```
    
    	-	Ensure all block math expressions are clearly separated from list items or inline text.
		-	Add a space after operators like \sum, \max, or any LaTeX commands followed by an underscore (_) to prevent Markdown parsers from interpreting _ as italic markers.  Mainly it is critical to put a space before "_". For example:

** Correct **
```
$$
R _{\text{BERT}} = \frac{\sum _{x _{i} \in x} \text{idf}(x _{i}) \cdot \max _{\hat{x} _{j} \in \hat{x}} x _{i^\top} \hat{x} _{j}}{\sum _{x _{i} \in x} \text{idf}(x _{i})}
$$
```

** Incorrect **
```
$$
R_{\text{BERT}} = \frac{\sum_{x_i \in x} \text{idf}(x_i) \cdot \max_{\hat{x}_j \in \hat{x}} x_i^\top \hat{x}_j}{\sum_{x_i \in x} \text{idf}(x_i)}
$$
```

	6. Citation: It is imperative that you do NOT make this up.  If the user does not explicitly provide the bibtex citation for the metric then you must say [More Information Needed].  If a citation is provided you must copy it EXACTLY.  Do NOT try to simplify any of the components such as the author list with an ellipsis.

## Categorical Suggestions for Consistency

Note: These suggestions are not exhaustive. While you should prioritize using the categories listed here for consistency, you may add new categories if the metric clearly warrants them.

### Domains

These represent broad areas of application for the metric. Choose one or more:
	-	Text Generation
	-	Speech Generation
	-	Code Generation
	-	Multimodal Generation
	-	Image Captioning
	-	Dialogue Systems
	-	Storytelling

### Tasks

These are specific tasks or use cases where the metric applies. Choose one or more:
	-	Machine Translation
	-	Summarization
	-	Paraphrasing
	-	Data-to-Text Generation
	-	Image-to-Text Generation
	-	Dialogue Generation
	-	Style Transfer
	-	Creative Writing (e.g., poetry, storytelling)
	-	Code Completion
	-	Response Generation

### Metric Type

These classify the metric based on its design and purpose. Choose one:
	-	Surface-Level Similarity (e.g., BLEU, ROUGE)
	-	Semantic Similarity (e.g., BERTScore)
	-	Fluency (e.g., perplexity-based metrics)
	-	Diversity (e.g., distinct-n)
	-	Robustness (e.g., adversarial robustness metrics)
	-	Fairness
	-	Faithfulness (e.g., factual consistency metrics)
	-	Reference-Free (e.g., coherence or novelty scoring)
	-	Explainability

### Inputs

These describe what the metric requires for evaluation:
	-	Reference-Based
	-	Reference-Free
	-	Input-Required
	-	Input-Optional

## Materials You Will Be Provided
	1.	Original Paper: The foundational paper introducing or defining the metric.
	2.	Reference Implementations (when available): Documentation from popular implementations (e.g., SacreBLEU README for BLEU).
	3.	Metric Card Template: The standardized structure for all Metric Cards (see below).
	4.	BLEU Metric Card Example: A high-quality example for reference.

=== TEMPLATE FOR METRIC CARDS ===
---
# Metric Card for {{ metric_name | default("Metric Name", true) }}

{{ metric_summary | default("A brief description of the metric and its purpose.", true) }}

## Metric Details

### Metric Description

{{ metric_description | default("Detailed explanation of the metric, including how it is calculated and what it measures.", true) }}

- **Metric Type:** {{ metric_type | default("[More Information Needed]", true) }}
- **Range:** {{ metric_range | default("[More Information Needed]", true) }}
- **Higher is Better?:** {{ higher_is_better | default("[More Information Needed]", true) }}
- **Reference-Based?:** {{ reference_based | default("[More Information Needed]", true) }}
- **Input-Required?:** {{ input_required | default("[More Information Needed]", true) }}

### Formal Definition

{{ metric_definition | default("Mathematical formula or detailed algorithmic definition.", true) }}

### Inputs and Outputs

- **Inputs:**  
  {{ metric_inputs | default("Description of required inputs (e.g., generated text, reference text, input prompt).", true) }}
  
- **Outputs:**  
  {{ metric_outputs | default("Description of the metric output (e.g., scalar score, distribution).", true) }}

## Intended Use

### Domains and Tasks

- **Domain:** {{ domain | default("[More Information Needed]", true) }}
- **Tasks:** {{ tasks | default("[More Information Needed]", true) }}

### Applicability and Limitations

- **Best Suited For:** {{ best_suited_for | default("[More Information Needed]", true) }}
- **Not Recommended For:** {{ not_recommended_for | default("[More Information Needed]", true) }}

## Metric Implementation

### Reference Implementations

- **Libraries/Packages:** {{ libraries | default("[More Information Needed]", true) }}

### Computational Complexity

- **Efficiency:** {{ efficiency | default("[More Information Needed]", true) }}
- **Scalability:** {{ scalability | default("[More Information Needed]", true) }}

## Known Limitations

{{ known_limitations | default("[More Information Needed]", true) }}

- **Biases:** {{ biases | default("Potential biases inherent in the metric.", true) }}
- **Task Misalignment Risks:** {{ task_misalignment | default("[More Information Needed]", true) }}
- **Failure Cases:** {{ failure_cases | default("[More Information Needed]", true) }}

## Related Metrics

{{ related_metrics | default("[More Information Needed]", true) }}

## Further Reading

- **Papers:** {{ papers | default("[More Information Needed]", true) }}
- **Blogs/Tutorials:** {{ blogs | default("[More Information Needed]", true) }}

## Citation

{{ bibtex_citation | default("[More Information Needed]", true) }}

## Metric Card Authors

- **Authors:** {{ metric_authors | default("[More Information Needed]", true) }}
- **Acknowledgment of AI Assistance:**
  {{ ai_assistance | default("Portions of this metric card were drafted with assistance from generative AI. All content has been reviewed and curated by the author to ensure accuracy.", true) }}  
- **Contact:** {{ metric_contact | default("[More Information Needed]", true) }}
======

=== BLEU Metric Card Example ===
---
# Metric Card for BLEU

BLEU (Bilingual Evaluation Understudy) is a widely used metric for evaluating the quality of text generated in tasks like machine translation and summarization. It measures the overlap of n-grams between a generated text and one or more reference texts, with a brevity penalty to penalize overly short translations. SacreBLEU, a modern implementation, ensures reproducibility and standardization of BLEU scores across research.

## Metric Details

### Metric Description

BLEU evaluates the quality of text generation by comparing n-grams in the generated output with those in one or more reference texts. It computes modified precision for n-grams and combines scores using a geometric mean, with a brevity penalty to ensure the length of the generated text matches that of the references. Higher BLEU scores indicate closer similarity to the references.

- **Metric Type:** Surface-Level Similarity
- **Range:** 0 to 1
- **Higher is Better?:** Yes
- **Reference-Based?:** Yes
- **Input-Required?:** No

### Formal Definition

$$
\text{BLEU} = \text{BP} \cdot \exp \left( \sum_{n=1}^N w_n \log p_n \right)
$$

where:
- $\text{BP} = \min(1, e^{1 - r/c})$ is the brevity penalty,
- $r$ is the effective reference length (based on the closest matching reference length for each sentence),
- $c$ is the candidate translation length,
- $p_n$ is the modified precision for n-grams of length $n$,
- $w_n$ are weights for each n-gram (commonly uniform, $w_n = \frac{1}{N}$).

### Inputs and Outputs

- **Inputs:**  
  - Generated text (candidate translation)  
  - Reference text(s) (gold-standard translations)  

- **Outputs:**  
  - Scalar BLEU score (range: 0 to 1)

## Intended Use

### Domains and Tasks

- **Domain:** Text Generation
- **Tasks:** Machine Translation, Summarization, Data-to-Text Generation

### Applicability and Limitations

- **Best Suited For:**  
  Structured tasks with a clear correspondence between generated and reference texts, such as translation or summarization.
  
- **Not Recommended For:**  
  Open-ended or creative generation tasks where diversity or semantic similarity matters more than lexical overlap (e.g., storytelling, dialogue).

## Metric Implementation

### Reference Implementations

- **Libraries/Packages:**
  - [SacreBLEU](https://github.com/mjpost/sacrebleu) (robust, standard implementation)
  - [NLTK](https://www.nltk.org/api/nltk.translate.html) (basic Python implementation)
  - [Hugging Face `evaluate`](https://huggingface.co/docs/evaluate) (integrated metric framework)

### Computational Complexity

- **Efficiency:**  
  BLEU is computationally efficient, requiring $O(n \cdot m)$ operations for $n$-gram matching where $n$ is the number of words in the candidate text and $m$ is the number of reference words. SacreBLEU optimizes tokenization and scoring, making it highly suitable for large-scale evaluations.

- **Scalability:**  
  BLEU scales well across datasets of varying sizes due to its simple design. SacreBLEU further supports evaluation with multiple references, diverse tokenization schemes, and language-specific preprocessing, making it adaptable to diverse evaluation setups.

## Known Limitations

- **Biases:**  
  - BLEU penalizes valid paraphrases or semantically equivalent outputs that do not match reference n-grams exactly.  
  - The brevity penalty can overly penalize valid shorter outputs, particularly for tasks where shorter text may be acceptable or even preferred (e.g., summarization).  

- **Task Misalignment Risks:**  
  - BLEU is not designed for evaluating tasks with high diversity in acceptable outputs (e.g., open-ended dialogue).  
  - Scores depend on the quality and number of references; fewer or inconsistent references can lead to misleading evaluations.

- **Failure Cases:**  
  - BLEU struggles to capture semantic adequacy beyond lexical similarity. For instance, it cannot identify whether a translation preserves the meaning of the original sentence if word choices diverge significantly.

## Related Metrics

- **ROUGE:** Often used for summarization tasks, emphasizing recall over precision.  
- **METEOR:** Incorporates synonym matching for better semantic alignment.  
- **BERTScore:** Uses contextual embeddings for semantic similarity.  

## Further Reading

- **Papers:**  
  - [Original BLEU Paper (Papineni et al., 2002)](https://www.aclweb.org/anthology/P02-1040)  
  - [SacreBLEU: A Call for Clarity in Reporting BLEU Scores (Post, 2018)](https://www.aclweb.org/anthology/W18-6319)
  
- **Blogs/Tutorials:**  
  - [Understanding BLEU](https://machinelearningmastery.com /calculate-bleu-score-for-text-python/)  
  - [SacreBLEU Documentation](https://github.com/mjpost/sacrebleu)

## Citation

@inproceedings{papineni-etal-2002-bleu,
    title = "{B}leu: a Method for Automatic Evaluation of Machine Translation",
    author = "Papineni, Kishore  and
      Roukos, Salim  and
      Ward, Todd  and
      Zhu, Wei-Jing",
    editor = "Isabelle, Pierre  and
      Charniak, Eugene  and
      Lin, Dekang",
    booktitle = "Proceedings of the 40th Annual Meeting of the Association for Computational Linguistics",
    month = jul,
    year = "2002",
    address = "Philadelphia, Pennsylvania, USA",
    publisher = "Association for Computational Linguistics",
    url = "https://aclanthology.org/P02-1040/",
    doi = "10.3115/1073083.1073135",
    pages = "311--318"
}

## Metric Card Authors

- **Authors:** Michael J. Ryan  
- **Acknowledgment of AI Assistance:**  
  Portions of this metric card were drafted with assistance from OpenAI's ChatGPT, based on user-provided inputs and relevant documentation. All content has been reviewed and curated by the author to ensure accuracy.  
- **Contact:** michaeljryan@stanford.edu
======

The metric you will be designing a card for is {Metric Name}

=== {SUPPLEMENTAL MATERIALS} ===

======

Now please write a high quality metric card for {Metric Name} given the provided materials!

Final **Important** Note: If the provided materials do not give enough information about a particular point for the metric (e.g. limitations or biases aren't listed) then do NOT make things up.  You can leave blanks or "Needs more information" where needed.  It is absolutely essential not to make things up or guess when producing this documentation otherwise future researchers and engineers will be confused and led astray.  Avoid making up links that you aren't fully confident in the url.

Remember to surround your answer in ```.  Thanks!
\end{lstlisting}
\end{PromptBox}

\subsection{DSPy Signatures}

\begin{DSPyBox}[Signature: GeneratePerturbationStrategies]
\textbf{Instruction:}
\begin{lstlisting}[style=dspy-instr]
You will be given:
- A Task description
- A Dimension to prioritize when perturbing outputs
- The Example Input, optional Example Reference, and Example Output

Instructions:
Your primary focus should be on degrading performance along the specified Dimension.
1. Begin with a rich reasoning paragraph (3-5 sentences) that explores a variety of ways to subtly degrade model outputs. Do not reference the specific example.
2. Under the heading **Strategies:**, list 1-3 numbered, high-level perturbation strategies.
   - Each strategy should be a short phrase (5-15 words) naming the category of change, followed by one concise sentence of abstract explanation.
   - Do not include concrete rewrites, instance-specific examples, or example sentences.
\end{lstlisting}

\medskip
\textbf{Inputs:}

\begin{DSPyFields}
\DSPyField{task}{str}{The task the model was originally trying to complete}
\DSPyField{example_sets}{list[str]}{Example inputs, outputs, and (optionally) references showing task performance}
\DSPyField{dimension}{str}{The dimension to prioritize for the perturbation}
\end{DSPyFields}

\medskip
\textbf{Outputs:}

\begin{DSPyFields}
\DSPyField{perturbation_ strategies}{list[str]}{1-3 high-level strategies to test robustness}
\end{DSPyFields}
\end{DSPyBox}

\begin{DSPyBox}[Signature: PerturbWorse]

\textbf{Instruction:}
\begin{lstlisting}[style=dspy-instr]
You will be given:  
    - A Task description  
    - A Dimension to prioritize when perturbing outputs  
    - The Example Input, optional Example Reference, and Model Output  
    - A perturbation_strength value ("subtle" or "obvious")  
    - A list of perturbation_strategies to apply  

Instructions:  
Your goal is to apply each strategy to the Model Output and produce a degraded version that specifically harms performance along the given Dimension, using the specified strength.  
Under the heading **Perturbed Outputs:**, return exactly one perturbed output per strategy.  
    - For **subtle** strength, introduce minimal distortion.  
    - For **obvious** strength, introduce more pronounced degradation.  
Do **not** include any reasoning, explanations, or examples -- only the perturbed text.
\end{lstlisting}

\medskip
\textbf{Inputs:}

\begin{DSPyFields}
\DSPyField{task}{str}{The task that the model was originally trying to complete}
\DSPyField{dimension}{str}{The dimension to prioritize for the perturbation (this should be the aspect of the model output that is most impacted by the perturbation)}
\DSPyField{input}{str}{The input provided to the model}
\DSPyField{references}{Union[list[str], None]}{The references of good outputs (may be None)}
\DSPyField{model_output}{str}{The output produced by the model}
\DSPyField{perturbation_ strength}{Literal['subtle', 'obvious']}{The strength of the perturbation (subtle or obvious)}
\DSPyField{perturbation_ strategies}{list[str]}{The perturbation strategies to use}
\end{DSPyFields}

\medskip
\textbf{Outputs:}

\begin{DSPyFields}
\DSPyField{perturbed_ outputs}{list[str]}{Perturbed text that is worse than the original model output.  Produce one perturbed output per strategy.}
\end{DSPyFields}
\end{DSPyBox}

\begin{DSPyBox}[Signature: PerturbSame]

\textbf{Instruction:}
\begin{lstlisting}[style=dspy-instr]
You will be given:
    - A Task description  
    - A Dimension to preserve when perturbing outputs  
    - The Example Input, optional Example Reference, and Model Output  
    - A perturbation_strength value ("subtle" or "obvious")  

Instructions:
Apply a perturbation to the Model Output that **maintains** performance on the specified Dimension.
Under the heading **Perturbed Output:** return exactly one string:
    - For **subtle** strength, apply a minimal change that does not impair the target Dimension.
    - For **obvious** strength, apply a more noticeable change that still keeps the target Dimension intact.
Some examples of types of perturbations would include: rephrasing, reordering, replacing words with synonyms, stylistic changes, etc. that do not impair the target Dimension.
If any change would harm the specified Dimension, simply return the original Model Output.
After producing your original plan/reasoning do **not** include any more reasoning, explanations, or examples -- only the perturbed text.
\end{lstlisting}

\medskip
\textbf{Inputs:}

\begin{DSPyFields}
\DSPyField{task}{str}{The task that the model was originally trying to complete}
\DSPyField{input}{str}{The input provided to the model}
\DSPyField{references}{Union[list[str], None]}{The references of good outputs (may be None)}
\DSPyField{model_output}{str}{The output produced by the model}
\DSPyField{perturbation_ strength}{Literal['subtle', 'obvious']}{The strength of the perturbation (subtle or obvious)}
\DSPyField{dimension}{str}{The aspect of the model output that MUST be preserved in quality}
\end{DSPyFields}

\medskip
\textbf{Outputs:}

\begin{DSPyFields}
\DSPyField{perturbed_output}{str}{Perturbed text that preserves performance along the given Dimension.}
\end{DSPyFields}
\end{DSPyBox}

\begin{DSPyBox}[Signature: LLMAsAJudgeSignature]

\textbf{Instruction:}
\begin{lstlisting}[style=dspy-instr]
Given an input text, the task description that the model was trying to follow, and a measure to rate the text on, return a score on this measure.
\end{lstlisting}

\medskip
\textbf{Inputs:}

\begin{DSPyFields}
\DSPyField{text}{Any}{The input text that we want to rate.}
\DSPyField{task_description}{Any}{A description of the task that the model was trying to solve when it generated the text. Could be left blank if not available.}
\DSPyField{measure}{Any}{The measure that we want to rate the text on.}
\DSPyField{suggested_range}{Any}{The suggested range of possible values for the measure.}
\end{DSPyFields}

\medskip
\textbf{Outputs:}

\begin{DSPyFields}
\DSPyField{score}{Any}{The score that the text should receive on this measure.}
\end{DSPyFields}
\end{DSPyBox}

\begin{DSPyBox}[Signature: LLMMetricRecommendationSignature]

\textbf{Instruction:}
\begin{lstlisting}[style=dspy-instr]
I am looking for a metric to evaluate the attached task. In particular I care about the specific target measurement that I attached.

Please help me decide from among the metrics that I have attached documentation for which one is most relevant to the task and target.

Please provide a ranking of the metrics from most relevant to least relevant for the task and target above.
You can reason first about what makes a metric relevant for the task and target, and then provide your ranking.

IMPORTANT: The final ranking should be a list of EXACT metric class names (no hyphens, no spaces, no extra words).  Use the METRIC NAME not what it is called in the documentation.
For example, use "SelfBLEU" not "Self-BLEU", use "BERTScore" not "BERT Score", use "BLEU" not "BLEU Score".

The final ranking should just be a list of metric names, in order from most relevant to least relevant.
The list should be exactly `num_metrics_to_recommend` items long.
\end{lstlisting}

\medskip
\textbf{Inputs:}

\begin{DSPyFields}
\DSPyField{task_description}{str}{A description of the task that an LLM performed and that I now want to evaluate.}
\DSPyField{target}{str}{The specific target measurement that I want to evaluate about the task.}
\DSPyField{metric_ documentation}{List[str]}{A list of metric names and their documentation.  The documentation will contain the metric name, as well as many details about the metric.}
\DSPyField{num_metrics_ to_recommend}{int}{The number of metrics to recommend.  It is imperative to target this number or very very close to it.  We will do more extensive filtering later.}
\end{DSPyFields}

\medskip
\textbf{Outputs:}

\begin{DSPyFields}
\DSPyField{ranking}{List[str]}{A numbered list of EXACT metric class names (no hyphens, no spaces, no extra words), in order from most relevant to least relevant. The list should be of length `num\_metrics\_to\_recommend`.  You should write the number in front of the metric name (e.g '1. METRIC1\_NAME', '2. METRIC2\_NAME', etc.).  REMEMBER: Put quotes around EACH number + metric name pair, not just one set of quotes for the full string.  IMPORTANT: Refer to "METRIC NAME: ..." for the exact name of the metric or it won't be a match.}
\end{DSPyFields}
\end{DSPyBox}

\begin{DSPyBox}[Signature: GenerateRubricSignature]

\textbf{Instruction:}
\begin{lstlisting}[style=dspy-instr]
Given a dataset, task description, and an evaluation metric, generate a rubric for the metric scoring from 1 to 5.
\end{lstlisting}

\medskip
\textbf{Inputs:}

\begin{DSPyFields}
\DSPyField{task_description}{Any}{A description of the task that the model is trying to solve.}
\DSPyField{good_examples}{Any}{A list of good examples of outputs for a model.}
\DSPyField{bad_examples}{Any}{A list of bad examples of outputs for a model.}
\DSPyField{metric_title}{Any}{The title of the metric.}
\DSPyField{metric_ description}{Any}{A description of the metric.}
\end{DSPyFields}

\medskip
\textbf{Outputs:}

\begin{DSPyFields}
\DSPyField{score_one_ description}{Any}{A description of what a score of 1 means.  This can be a bullet point list of what criteria to look for in assigning a score of 1.}
\DSPyField{score_two_ description}{Any}{A description of what a score of 2 means.  This can be a bullet point list of what criteria to look for in assigning a score of 2.}
\DSPyField{score_three_ description}{Any}{A description of what a score of 3 means.  This can be a bullet point list of what criteria to look for in assigning a score of 3.}
\DSPyField{score_four_ description}{Any}{A description of what a score of 4 means.  This can be a bullet point list of what criteria to look for in assigning a score of 4.}
\DSPyField{score_five_ description}{Any}{A description of what a score of 5 means.  This can be a bullet point list of what criteria to look for in assigning a score of 5.}
\end{DSPyFields}
\end{DSPyBox}

\begin{DSPyBox}[Signature: GenerateAxisOfVariationSignature]

\textbf{Instruction:}
\begin{lstlisting}[style=dspy-instr]
Given a task description, a target metric, and good/bad examples, generate a list of axes of variation which could be used to explain the differences between the good and bad examples.  These axes of variation will be used as measures to evaluate the model's performance, so they should be informative and useful for the model to improve on.
\end{lstlisting}

\medskip
\textbf{Inputs:}

\begin{DSPyFields}
\DSPyField{task_description}{str}{A description of the overall task the model is trying to solve.}
\DSPyField{target_name}{Optional[str]}{Optional hint of the target metric/column we care about. Could be 'None' or something generic like 'quality' or 'score'.}
\DSPyField{good_examples}{List[str]}{A list of examples with *high* quality according to the target metric.}
\DSPyField{bad_examples}{List[str]}{A list of examples with *low* quality according to the target metric.}
\DSPyField{num_axes_to _generate}{int}{The number of axes of variation to generate.}
\end{DSPyFields}

\medskip
\textbf{Outputs:}

\begin{DSPyFields}
\DSPyField{axes_of _variation}{List[str]}{An ordered list (most-important first) describing possible axes of variation. Please bold the name of the axis of variation (e.g. **Axes Name**), and ALSO include a brief sentence-long explanation of the axis of variation. (e.g. **Axes Name** Brief Explanation).  Please include exactly 'num\_axes\_to\_generate' axes of variation in the output.  Avoid special characters since they sometimes mess up the parsing.}
\end{DSPyFields}
\end{DSPyBox}

\section{Example Metric Card: BLEU}
\label{appendix:example_metric_card}

\noindent
\textbf{Metric Card for BLEU} \\

\noindent
BLEU (Bilingual Evaluation Understudy) is a widely used metric for evaluating the quality of text generated in tasks like machine translation and summarization. It measures the overlap of n-grams between a generated text and one or more reference texts, with a brevity penalty to penalize overly short translations. SacreBLEU, a modern implementation, ensures reproducibility and standardization of BLEU scores across research.

\vspace{1em}
\noindent
\textbf{Metric Details} \\

\noindent
\textit{Metric Description} \\
BLEU evaluates the quality of text generation by comparing n-grams in the generated output with those in one or more reference texts. It computes modified precision for n-grams and combines scores using a geometric mean, with a brevity penalty to ensure the length of the generated text matches that of the references. Higher BLEU scores indicate closer similarity to the references.

\begin{itemize}
    \item \textbf{Metric Type:} Surface-Level Similarity
    \item \textbf{Range:} 0 to 1
    \item \textbf{Higher is Better?:} Yes
    \item \textbf{Reference-Based?:} Yes
    \item \textbf{Input-Required?:} No
\end{itemize}

\noindent
\textit{Formal Definition} \\
\[
\text{BLEU} = \text{BP} \cdot \exp \left( \sum_{n=1}^N w_n \log p_n \right)
\]
\begin{itemize}
    \item $\text{BP} = \min(1, e^{1 - r/c})$ is the brevity penalty,
    \item $r$ is the effective reference length (based on the closest matching reference length for each sentence),
    \item $c$ is the candidate translation length,
    \item $p_n$ is the modified precision for n-grams of length $n$,
    \item $w_n$ are weights for each n-gram (commonly uniform, $w_n = \frac{1}{N}$).
\end{itemize}

\noindent
\textit{Inputs and Outputs}
\begin{itemize}
    \item \textbf{Inputs:}
    \begin{itemize}
        \item Generated text (candidate translation)
        \item Reference text(s) (gold-standard translations)
    \end{itemize}
    \item \textbf{Outputs:}
    \begin{itemize}
        \item Scalar BLEU score (range: 0 to 1)
    \end{itemize}
\end{itemize}

\vspace{1em}
\noindent
\textbf{Intended Use} \\

\noindent
\textit{Domains and Tasks}
\begin{itemize}
    \item \textbf{Domain:} Text Generation
    \item \textbf{Tasks:} Machine Translation, Summarization, Data-to-Text Generation
\end{itemize}

\noindent
\textit{Applicability and Limitations}
\begin{itemize}
    \item \textbf{Best Suited For:} Structured tasks with a clear correspondence between generated and reference texts, such as translation or summarization.
    \item \textbf{Not Recommended For:} Open-ended or creative generation tasks where diversity or semantic similarity matters more than lexical overlap (e.g., storytelling, dialogue).
\end{itemize}

\vspace{1em}
\noindent
\textbf{Metric Implementation} \\

\noindent
\textit{Reference Implementations}
\begin{itemize}
    \item \textbf{Libraries/Packages:}
    \begin{itemize}
        \item \href{https://github.com/mjpost/sacrebleu}{SacreBLEU} (robust, standard implementation)
        \item \href{https://www.nltk.org/api/nltk.translate.html}{NLTK} (basic Python implementation)
        \item \href{https://huggingface.co/docs/evaluate}{Hugging Face \texttt{evaluate}} (integrated metric framework)
    \end{itemize}
\end{itemize}

\noindent
\textit{Computational Complexity}
\begin{itemize}
    \item \textbf{Efficiency:} BLEU is computationally efficient, requiring $O(n \cdot m)$ operations for $n$-gram matching where $n$ is the number of words in the candidate text and $m$ is the number of reference words. SacreBLEU optimizes tokenization and scoring, making it highly suitable for large-scale evaluations.
    \item \textbf{Scalability:} BLEU scales well across datasets of varying sizes due to its simple design. SacreBLEU further supports evaluation with multiple references, diverse tokenization schemes, and language-specific preprocessing, making it adaptable to diverse evaluation setups.
\end{itemize}

\vspace{1em}
\noindent
\textbf{Known Limitations} \\
\begin{itemize}
    \item \textbf{Biases:}
    \begin{itemize}
        \item BLEU penalizes valid paraphrases or semantically equivalent outputs that do not match reference n-grams exactly.
        \item The brevity penalty can overly penalize valid shorter outputs, particularly for tasks where shorter text may be acceptable or even preferred (e.g., summarization).
    \end{itemize}
    \item \textbf{Task Misalignment Risks:}
    \begin{itemize}
        \item BLEU is not designed for evaluating tasks with high diversity in acceptable outputs (e.g., open-ended dialogue).
        \item Scores depend on the quality and number of references; fewer or inconsistent references can lead to misleading evaluations.
    \end{itemize}
    \item \textbf{Failure Cases:}
    \begin{itemize}
        \item BLEU struggles to capture semantic adequacy beyond lexical similarity. For instance, it cannot identify whether a translation preserves the meaning of the original sentence if word choices diverge significantly.
    \end{itemize}
\end{itemize}

\vspace{1em}
\noindent
\textbf{Related Metrics} \\
\begin{itemize}
    \item \textbf{ROUGE:} Often used for summarization tasks, emphasizing recall over precision.
    \item \textbf{METEOR:} Incorporates synonym matching for better semantic alignment.
    \item \textbf{BERTScore:} Uses contextual embeddings for semantic similarity.
\end{itemize}

\vspace{1em}
\noindent
\textbf{Further Reading} \\
\begin{itemize}
    \item \textbf{Papers:}
    \begin{itemize}
        \item \href{https://www.aclweb.org/anthology/P02-1040}{Original BLEU Paper (Papineni et al., 2002)}
        \item \href{https://www.aclweb.org/anthology/W18-6319}{SacreBLEU: A Call for Clarity in Reporting BLEU Scores (Post, 2018)}
    \end{itemize}
    \item \textbf{Blogs/Tutorials:}
    \begin{itemize}
        \item \href{https://machinelearningmastery.com/calculate-bleu-score-for-text-python/}{Understanding BLEU}
        \item \href{https://github.com/mjpost/sacrebleu}{SacreBLEU Documentation}
    \end{itemize}
\end{itemize}

\vspace{1em}
\noindent
\textbf{Metric Card Authors} \\
\begin{itemize}
    \item \textbf{Authors:} Michael J. Ryan
    \item \textbf{Acknowledgment of AI Assistance:} Portions of this metric card were drafted with assistance from OpenAI's ChatGPT, based on user-provided inputs and relevant documentation. All content has been reviewed and curated by the author to ensure accuracy.
    \item \textbf{Contact:} michaeljryan@stanford.edu
\end{itemize}

\section{AutoMetrics Design Ablations}
\label{appendix:autometrics_design_ablations}

\subsection{Retrieve}
\label{appendix:subsec:retrieve}

For our retrieval experiments we run all metrics in the MetricBank to get the ground truth kendall correlation on the development set.  With this we know the true rank order of the metrics.  We then perform retrieval using a set of retrieval algorithms, namely BM25, ColBERT, Faiss, and using an LLM with all documents in context.  We additionally try pipelined versions of all of these retrievers feeding into an LLM.  We report Recall@[1,5,10,20] and NDCG@[1,5,10,20] in Table~\ref{tab:recmetricskendallqwen3} for Qwen3-32B and Table~\ref{tab:recmetricskendallgpt4omini} for GPT-4o-mini.

\begin{table*}[h]
  \centering
  \resizebox{\textwidth}{!}{%
  \begin{tabular}{lcccc|cccc}
    \toprule
    \rowcolor[gray]{0.85}
    & \multicolumn{4}{c|}{\textbf{NDCG}} &
      \multicolumn{4}{c}{\textbf{Recall}}\\
    \rowcolor[gray]{0.85}
    Method & @1 & @5 & @10 & @20 & @1 & @5 & @10 & @20\\
    \midrule
    BM25 & 0.208 {\scriptsize $\pm$ 0.274} & 0.342 {\scriptsize $\pm$ 0.171} & 0.427 {\scriptsize $\pm$ 0.143} & 0.567 {\scriptsize $\pm$ 0.16} & 0.065 {\scriptsize $\pm$ 0.095} & 0.224 {\scriptsize $\pm$ 0.146} & 0.418 {\scriptsize $\pm$ 0.173} & \textbf{0.788 {\scriptsize $\pm$ 0.319}}\\
    ColBERT & 0.272 {\scriptsize $\pm$ 0.293} & 0.343 {\scriptsize $\pm$ 0.203} & 0.442 {\scriptsize $\pm$ 0.178} & 0.57 {\scriptsize $\pm$ 0.174} & 0.059 {\scriptsize $\pm$ 0.092} & 0.212 {\scriptsize $\pm$ 0.155} & 0.441 {\scriptsize $\pm$ 0.273} & 0.776 {\scriptsize $\pm$ 0.361}\\
    Faiss & 0.103 {\scriptsize $\pm$ 0.059} & 0.227 {\scriptsize $\pm$ 0.144} & 0.326 {\scriptsize $\pm$ 0.163} & 0.461 {\scriptsize $\pm$ 0.199} & 0.018 {\scriptsize $\pm$ 0.058} & 0.171 {\scriptsize $\pm$ 0.204} & 0.353 {\scriptsize $\pm$ 0.256} & 0.694 {\scriptsize $\pm$ 0.427}\\
    LLMRec & 0.31 {\scriptsize $\pm$ 0.334} & 0.396 {\scriptsize $\pm$ 0.249} & 0.478 {\scriptsize $\pm$ 0.219} & 0.602 {\scriptsize $\pm$ 0.196} & 0.088 {\scriptsize $\pm$ 0.101} & 0.294 {\scriptsize $\pm$ 0.204} & 0.465 {\scriptsize $\pm$ 0.273} & 0.641 {\scriptsize $\pm$ 0.264}\\
    BM25→LLMRec & 0.316 {\scriptsize $\pm$ 0.323} & 0.42 {\scriptsize $\pm$ 0.226} & 0.498 {\scriptsize $\pm$ 0.197} & 0.603 {\scriptsize $\pm$ 0.186} & \textbf{0.094 {\scriptsize $\pm$ 0.101}} & 0.312 {\scriptsize $\pm$ 0.179} & 0.494 {\scriptsize $\pm$ 0.257} & 0.665 {\scriptsize $\pm$ 0.245}\\
    ColBERT→LLMRec & \textbf{0.403 {\scriptsize $\pm$ 0.387}} & \textbf{0.462 {\scriptsize $\pm$ 0.28}} & \textbf{0.528 {\scriptsize $\pm$ 0.238}} & \textbf{0.631 {\scriptsize $\pm$ 0.212}} & \textbf{0.094 {\scriptsize $\pm$ 0.101}} & \textbf{0.329 {\scriptsize $\pm$ 0.225}} & \textbf{0.518 {\scriptsize $\pm$ 0.266}} & 0.694 {\scriptsize $\pm$ 0.21}\\
    Faiss→LLMRec & 0.164 {\scriptsize $\pm$ 0.186} & 0.324 {\scriptsize $\pm$ 0.234} & 0.393 {\scriptsize $\pm$ 0.215} & 0.529 {\scriptsize $\pm$ 0.216} & 0.065 {\scriptsize $\pm$ 0.095} & 0.247 {\scriptsize $\pm$ 0.226} & 0.4 {\scriptsize $\pm$ 0.27} & 0.6 {\scriptsize $\pm$ 0.304}\\
    \bottomrule
  \end{tabular}}
  \caption{Average performance (± std) across all tasks/axes using Kendall ground truth (recommendations from \texttt{qwen3}).}
  \label{tab:recmetricskendallqwen3}
\end{table*}

\begin{table*}[h]
  \centering
  \resizebox{\textwidth}{!}{%
  \begin{tabular}{lcccc|cccc}
    \toprule
    \rowcolor[gray]{0.85}
    & \multicolumn{4}{c|}{\textbf{NDCG}} &
      \multicolumn{4}{c}{\textbf{Recall}}\\
    \rowcolor[gray]{0.85}
    Method & @1 & @5 & @10 & @20 & @1 & @5 & @10 & @20\\
    \midrule
    BM25 & 0.208 {\scriptsize $\pm$ 0.274} & 0.342 {\scriptsize $\pm$ 0.171} & 0.427 {\scriptsize $\pm$ 0.143} & 0.567 {\scriptsize $\pm$ 0.16} & 0.065 {\scriptsize $\pm$ 0.095} & 0.224 {\scriptsize $\pm$ 0.146} & 0.418 {\scriptsize $\pm$ 0.173} & 0.788 {\scriptsize $\pm$ 0.319}\\
    ColBERT & 0.272 {\scriptsize $\pm$ 0.293} & 0.343 {\scriptsize $\pm$ 0.201} & 0.42 {\scriptsize $\pm$ 0.179} & 0.568 {\scriptsize $\pm$ 0.167} & 0.059 {\scriptsize $\pm$ 0.092} & 0.247 {\scriptsize $\pm$ 0.191} & 0.429 {\scriptsize $\pm$ 0.27} & \textbf{0.8 {\scriptsize $\pm$ 0.338}}\\
    Faiss & 0.098 {\scriptsize $\pm$ 0.059} & 0.21 {\scriptsize $\pm$ 0.128} & 0.314 {\scriptsize $\pm$ 0.167} & 0.461 {\scriptsize $\pm$ 0.19} & 0.012 {\scriptsize $\pm$ 0.048} & 0.159 {\scriptsize $\pm$ 0.169} & 0.371 {\scriptsize $\pm$ 0.304} & 0.729 {\scriptsize $\pm$ 0.378}\\
    LLMRec & 0.261 {\scriptsize $\pm$ 0.302} & 0.416 {\scriptsize $\pm$ 0.249} & 0.502 {\scriptsize $\pm$ 0.235} & 0.585 {\scriptsize $\pm$ 0.217} & 0.076 {\scriptsize $\pm$ 0.099} & 0.347 {\scriptsize $\pm$ 0.243} & 0.518 {\scriptsize $\pm$ 0.316} & 0.759 {\scriptsize $\pm$ 0.326}\\
    BM25→LLMRec & 0.206 {\scriptsize $\pm$ 0.197} & 0.394 {\scriptsize $\pm$ 0.196} & 0.47 {\scriptsize $\pm$ 0.175} & 0.576 {\scriptsize $\pm$ 0.162} & 0.076 {\scriptsize $\pm$ 0.099} & 0.347 {\scriptsize $\pm$ 0.233} & 0.512 {\scriptsize $\pm$ 0.257} & 0.794 {\scriptsize $\pm$ 0.297}\\
    ColBERT→LLMRec & \textbf{0.328 {\scriptsize $\pm$ 0.31}} & \textbf{0.475 {\scriptsize $\pm$ 0.251}} & \textbf{0.55 {\scriptsize $\pm$ 0.214}} & \textbf{0.628 {\scriptsize $\pm$ 0.198}} & \textbf{0.1 {\scriptsize $\pm$ 0.102}} & \textbf{0.388 {\scriptsize $\pm$ 0.246}} & \textbf{0.565 {\scriptsize $\pm$ 0.301}} & 0.759 {\scriptsize $\pm$ 0.333}\\
    Faiss→LLMRec & 0.157 {\scriptsize $\pm$ 0.124} & 0.325 {\scriptsize $\pm$ 0.205} & 0.406 {\scriptsize $\pm$ 0.201} & 0.526 {\scriptsize $\pm$ 0.212} & 0.065 {\scriptsize $\pm$ 0.095} & 0.276 {\scriptsize $\pm$ 0.226} & 0.424 {\scriptsize $\pm$ 0.31} & 0.635 {\scriptsize $\pm$ 0.37}\\
    \bottomrule
  \end{tabular}}
  \caption{Average performance (± std) across all tasks/axes using Kendall ground truth (recommendations from \texttt{gpt4o-mini}).}
  \label{tab:recmetricskendallgpt4omini}
\end{table*}

Overall we find that ColBERT $\rightarrow$ LLMRec is consistently the best approach for retrieval, performing the best across 14/16 of our evaluation settings.  Thus, we use ColBERT $\rightarrow$ LLMRec for all metric retrieval in the main paper.

\subsection{Generate}
\label{appendix:subsec:generate}

We test eight different approaches to metric generation.  Of these approaches five of them are cheap to produce, while three of them are expensive/time-consuming to produce.

For \textbf{CodeGen} we prompt an LLM to propose “axes of variation” from high/low examples and then synthesize small, executable Python snippets that implement a scoring function (\texttt{compute\_score}). The generated code is cleaned, validated on a sample, and—if it errors—automatically repaired once by an LLM. We support both reference-free and reference-based variants.

For \textbf{G-Eval} \citep{liu-etal-2023-g} we convert each axis into a concrete evaluation criterion, auto-generate numbered evaluation steps, and prompt an LLM judge to produce a brief rationale followed by a discrete score (1--5). We request token-level log probabilities and, at the final score position (found by scanning backward), extract the logprobs over tokens $\{1,2,3,4,5\}$, softmax-normalize, and return the probability-weighted expectation $\hat{s}=\sum_{s=1}^{5} s\,P(s\,|\,\text{prompt},\text{rationale})$. Both reference-free and reference-based variants are supported.

For \textbf{Single Criteria} \citep{saadfalcon2024lmunitfinegrainedevaluationnatural} LLM-as-a-Judge, we show high-scoring and low-scoring data points to an LLM and ask for ``axes of variation." Each axis becomes its own metric, with the LLM prompted to output an integer score from 1--5. 

For \textbf{Rubric} \citep{gunjal2025rubricsrewardsreinforcementlearning} we add an additional step to Single Criteria where we ask an LLM to generate explanations of what 1--5 scores should contain for each rubric item.  

For \textbf{Rubric (Prometheus)} \citep{kim2024prometheus} we first synthesize a five-level rubric (descriptions for scores 1--5) from dataset examples, then use a Prometheus evaluator (e.g., M-Prometheus-14B) to assign scores conditioned on that rubric. This keeps the rubric explicit while using a strong, specialized judge.

\textbf{Finetune} is our first expensive to produce metric. For this we fine-tune a ModernBERT-Large regression head (with LoRA/PEFT) on formatted input–output (and references when available) to directly predict the target score. We use an 80/20 train/validation split, optimize with AdamW, and save the resulting adapter as a learned metric that runs without an LLM at inference.

For \textbf{Examples} we separate the provided human-rated examples into quintiles. Based on the context length of the LLM judge we determine how many examples we can reasonably sample from each quintile without exceeding the context length.  We try 5 randomly sampled sets of uniformly distributed examples as context in an LLM-judge prompt and select the set that minimizes average distance to human labels on the trainset. 

For \textbf{Prompt Optimization (MIPROv2)} we run DSPy's MIPROv2 \citep{opsahl-ong-etal-2024-optimizing} optimizer with \texttt{auto\_mode="medium"} on the provided data to generate informative examples and rewrite the evaluation prompt for an LLM judge. 

\begin{table*}[h]
  \centering
  \small
  \setlength{\tabcolsep}{6pt}
  \renewcommand{\arraystretch}{1.1}
  \resizebox{\textwidth}{!}{%
  \begin{tabular}{lccccc|ccc}
    \toprule
    \rowcolor{gray!30}
    & \multicolumn{5}{c|}{\textbf{Cheap to Produce}} & \multicolumn{3}{c}{\textbf{Expensive to Produce}} \\
    \rowcolor{gray!30}
    Task (Measure) & Code Gen & G-Eval & Single Criterion & Rubric (DSPy) & Rubric (Prometheus) & Finetune & Examples & MIPROv2 \\
    \midrule
        \rowcolor{gray!10}
        \multicolumn{9}{l}{\textit{\textbf{In-Distribution Tasks}: some metrics in our bank were designed to directly evaluate these tasks.}} \\
        \midrule
        SummEval (coherence) & 0.098 {\scriptsize $\pm$ 0.019} & 0.105 {\scriptsize $\pm$ 0.023} & \textbf{0.194 {\scriptsize $\pm$ 0.010}} & 0.173 {\scriptsize $\pm$ 0.017} & 0.140 {\scriptsize $\pm$ 0.016} & 0.104 {\scriptsize $\pm$ 0.016} & 0.226 {\scriptsize $\pm$ 0.019} & \textbf{0.227 {\scriptsize $\pm$ 0.044}} \\
        SummEval (consistency) & 0.083 {\scriptsize $\pm$ 0.018} & 0.102 {\scriptsize $\pm$ 0.013} & \textbf{0.173 {\scriptsize $\pm$ 0.023}} & 0.160 {\scriptsize $\pm$ 0.026} & 0.122 {\scriptsize $\pm$ 0.015} & 0.095 {\scriptsize $\pm$ 0.042} & \textbf{0.226 {\scriptsize $\pm$ 0.066}} & 0.199 {\scriptsize $\pm$ 0.030} \\
        SummEval (fluency) & 0.057 {\scriptsize $\pm$ 0.016} & 0.076 {\scriptsize $\pm$ 0.011} & \textbf{0.121 {\scriptsize $\pm$ 0.009}} & 0.110 {\scriptsize $\pm$ 0.015} & 0.096 {\scriptsize $\pm$ 0.017} & 0.061 {\scriptsize $\pm$ 0.016} & \textbf{0.146 {\scriptsize $\pm$ 0.015}} & 0.136 {\scriptsize $\pm$ 0.048} \\
        SummEval (relevance) & 0.097 {\scriptsize $\pm$ 0.025} & 0.144 {\scriptsize $\pm$ 0.026} & \textbf{0.213 {\scriptsize $\pm$ 0.017}} & 0.189 {\scriptsize $\pm$ 0.018} & 0.151 {\scriptsize $\pm$ 0.014} & 0.067 {\scriptsize $\pm$ 0.043} & 0.243 {\scriptsize $\pm$ 0.022} & \textbf{0.263 {\scriptsize $\pm$ 0.022}} \\
        Primock57 (inc\_plus\_omi) & 0.105 {\scriptsize $\pm$ 0.036} & 0.086 {\scriptsize $\pm$ 0.017} & \textbf{0.247 {\scriptsize $\pm$ 0.031}} & 0.188 {\scriptsize $\pm$ 0.043} & 0.196 {\scriptsize $\pm$ 0.025} & 0.090 {\scriptsize $\pm$ 0.057} & 0.253 {\scriptsize $\pm$ 0.057} & \textbf{0.258 {\scriptsize $\pm$ 0.067}} \\
        Primock57 (incorrect) & 0.145 {\scriptsize $\pm$ 0.073} & 0.060 {\scriptsize $\pm$ 0.014} & \textbf{0.250 {\scriptsize $\pm$ 0.059}} & 0.169 {\scriptsize $\pm$ 0.070} & 0.202 {\scriptsize $\pm$ 0.026} & 0.026 {\scriptsize $\pm$ 0.029} & \textbf{0.266 {\scriptsize $\pm$ 0.039}} & 0.213 {\scriptsize $\pm$ 0.164} \\
        Primock57 (omissions) & 0.123 {\scriptsize $\pm$ 0.059} & 0.061 {\scriptsize $\pm$ 0.021} & 0.119 {\scriptsize $\pm$ 0.029} & 0.116 {\scriptsize $\pm$ 0.025} & \textbf{0.129 {\scriptsize $\pm$ 0.020}} & 0.125 {\scriptsize $\pm$ 0.077} & \textbf{0.169 {\scriptsize $\pm$ 0.023}} & 0.122 {\scriptsize $\pm$ 0.097} \\
        Primock57 (time\_sec) & 0.102 {\scriptsize $\pm$ 0.038} & 0.053 {\scriptsize $\pm$ 0.009} & \textbf{0.159 {\scriptsize $\pm$ 0.026}} & 0.132 {\scriptsize $\pm$ 0.016} & --- & 0.058 {\scriptsize $\pm$ 0.049} & 0.057 {\scriptsize $\pm$ 0.050} & \textbf{0.129 {\scriptsize $\pm$ 0.041}} \\
        SimpEval (score) & 0.100 {\scriptsize $\pm$ 0.037} & 0.184 {\scriptsize $\pm$ 0.028} & \textbf{0.229 {\scriptsize $\pm$ 0.019}} & 0.192 {\scriptsize $\pm$ 0.012} & 0.155 {\scriptsize $\pm$ 0.017} & 0.046 {\scriptsize $\pm$ 0.037} & 0.216 {\scriptsize $\pm$ 0.036} & \textbf{0.243 {\scriptsize $\pm$ 0.130}} \\
        SimpDA (fluency) & 0.180 {\scriptsize $\pm$ 0.013} & 0.364 {\scriptsize $\pm$ 0.022} & \textbf{0.521 {\scriptsize $\pm$ 0.014}} & 0.511 {\scriptsize $\pm$ 0.018} & 0.460 {\scriptsize $\pm$ 0.025} & 0.050 {\scriptsize $\pm$ 0.051} & 0.582 {\scriptsize $\pm$ 0.017} & \textbf{0.583 {\scriptsize $\pm$ 0.058}} \\
        SimpDA (meaning) & 0.252 {\scriptsize $\pm$ 0.066} & 0.397 {\scriptsize $\pm$ 0.030} & \textbf{0.590 {\scriptsize $\pm$ 0.016}} & 0.570 {\scriptsize $\pm$ 0.020} & 0.546 {\scriptsize $\pm$ 0.026} & 0.055 {\scriptsize $\pm$ 0.038} & \textbf{0.632 {\scriptsize $\pm$ 0.025}} & 0.625 {\scriptsize $\pm$ 0.024} \\
        SimpDA (simplicity) & 0.173 {\scriptsize $\pm$ 0.024} & 0.305 {\scriptsize $\pm$ 0.033} & \textbf{0.523 {\scriptsize $\pm$ 0.030}} & 0.481 {\scriptsize $\pm$ 0.021} & 0.442 {\scriptsize $\pm$ 0.015} & 0.041 {\scriptsize $\pm$ 0.058} & 0.584 {\scriptsize $\pm$ 0.025} & \textbf{0.628 {\scriptsize $\pm$ 0.035}} \\
        HelpSteer (coherence) & 0.029 {\scriptsize $\pm$ 0.004} & 0.162 {\scriptsize $\pm$ 0.027} & \textbf{0.229 {\scriptsize $\pm$ 0.013}} & 0.190 {\scriptsize $\pm$ 0.013} & --- & 0.014 {\scriptsize $\pm$ 0.009} & 0.297 {\scriptsize $\pm$ 0.023} & \textbf{0.297 {\scriptsize $\pm$ 0.006}} \\
        HelpSteer (complexity) & \textbf{0.223 {\scriptsize $\pm$ 0.083}} & 0.122 {\scriptsize $\pm$ 0.029} & 0.149 {\scriptsize $\pm$ 0.042} & 0.184 {\scriptsize $\pm$ 0.035} & --- & 0.071 {\scriptsize $\pm$ 0.070} & \textbf{0.221 {\scriptsize $\pm$ 0.050}} & 0.095 {\scriptsize $\pm$ 0.018} \\
        HelpSteer (correctness) & 0.068 {\scriptsize $\pm$ 0.007} & 0.270 {\scriptsize $\pm$ 0.024} & \textbf{0.356 {\scriptsize $\pm$ 0.024}} & 0.342 {\scriptsize $\pm$ 0.018} & --- & 0.044 {\scriptsize $\pm$ 0.027} & 0.392 {\scriptsize $\pm$ 0.027} & \textbf{0.424 {\scriptsize $\pm$ 0.009}} \\
        HelpSteer (helpfulness) & 0.066 {\scriptsize $\pm$ 0.019} & 0.241 {\scriptsize $\pm$ 0.018} & \textbf{0.333 {\scriptsize $\pm$ 0.011}} & 0.327 {\scriptsize $\pm$ 0.018} & --- & 0.049 {\scriptsize $\pm$ 0.030} & \textbf{0.407 {\scriptsize $\pm$ 0.016}} & 0.402 {\scriptsize $\pm$ 0.013} \\
        HelpSteer (verbosity) & \textbf{0.290 {\scriptsize $\pm$ 0.028}} & 0.154 {\scriptsize $\pm$ 0.043} & 0.193 {\scriptsize $\pm$ 0.051} & 0.252 {\scriptsize $\pm$ 0.053} & --- & 0.084 {\scriptsize $\pm$ 0.031} & \textbf{0.406 {\scriptsize $\pm$ 0.015}} & 0.103 {\scriptsize $\pm$ 0.028} \\
        HelpSteer2 (coherence) & 0.024 {\scriptsize $\pm$ 0.005} & 0.116 {\scriptsize $\pm$ 0.019} & \textbf{0.154 {\scriptsize $\pm$ 0.005}} & 0.138 {\scriptsize $\pm$ 0.020} & --- & 0.043 {\scriptsize $\pm$ 0.032} & \textbf{0.192 {\scriptsize $\pm$ 0.016}} & 0.169 {\scriptsize $\pm$ 0.028} \\
        HelpSteer2 (complexity) & \textbf{0.113 {\scriptsize $\pm$ 0.040}} & 0.074 {\scriptsize $\pm$ 0.024} & 0.091 {\scriptsize $\pm$ 0.013} & 0.100 {\scriptsize $\pm$ 0.014} & --- & 0.096 {\scriptsize $\pm$ 0.000} & \textbf{0.335 {\scriptsize $\pm$ 0.074}} & 0.065 {\scriptsize $\pm$ 0.045} \\
        HelpSteer2 (correctness) & 0.052 {\scriptsize $\pm$ 0.012} & 0.167 {\scriptsize $\pm$ 0.012} & \textbf{0.245 {\scriptsize $\pm$ 0.007}} & 0.212 {\scriptsize $\pm$ 0.016} & --- & 0.037 {\scriptsize $\pm$ 0.035} & \textbf{0.332 {\scriptsize $\pm$ 0.017}} & 0.320 {\scriptsize $\pm$ 0.019} \\
        HelpSteer2 (helpfulness) & 0.068 {\scriptsize $\pm$ 0.009} & 0.134 {\scriptsize $\pm$ 0.015} & \textbf{0.217 {\scriptsize $\pm$ 0.019}} & 0.183 {\scriptsize $\pm$ 0.008} & 0.135 {\scriptsize $\pm$ 0.008} & 0.026 {\scriptsize $\pm$ 0.015} & 0.293 {\scriptsize $\pm$ 0.020} & \textbf{0.309 {\scriptsize $\pm$ 0.015}} \\
        HelpSteer2 (verbosity) & 0.224 {\scriptsize $\pm$ 0.018} & 0.161 {\scriptsize $\pm$ 0.031} & 0.210 {\scriptsize $\pm$ 0.052} & \textbf{0.234 {\scriptsize $\pm$ 0.048}} & --- & 0.167 {\scriptsize $\pm$ 0.068} & \textbf{0.432 {\scriptsize $\pm$ 0.015}} & 0.081 {\scriptsize $\pm$ 0.315} \\
        \midrule
        \rowcolor{gray!10}
        \multicolumn{9}{l}{\textit{\textbf{Out-of-Distribution Tasks}: no metric is specifically designed for these -- tests generalization and metric generation.}} \\
        \midrule
        EvalGenProduct (grade) & 0.262 {\scriptsize $\pm$ 0.046} & 0.285 {\scriptsize $\pm$ 0.029} & \textbf{0.343 {\scriptsize $\pm$ 0.085}} & 0.303 {\scriptsize $\pm$ 0.072} & 0.201 {\scriptsize $\pm$ 0.021} & 0.210 {\scriptsize $\pm$ 0.236} & 0.145 {\scriptsize $\pm$ 0.046} & \textbf{0.216 {\scriptsize $\pm$ 0.173}} \\
        EvalGenMedical (grade) & 0.262 {\scriptsize $\pm$ 0.046} & 0.285 {\scriptsize $\pm$ 0.029} & \textbf{0.343 {\scriptsize $\pm$ 0.085}} & 0.303 {\scriptsize $\pm$ 0.072} & 0.201 {\scriptsize $\pm$ 0.021} & 0.210 {\scriptsize $\pm$ 0.236} & 0.145 {\scriptsize $\pm$ 0.046} & \textbf{0.216 {\scriptsize $\pm$ 0.173}} \\
        RealHumanEval (accepted) & 0.046 {\scriptsize $\pm$ 0.007} & 0.039 {\scriptsize $\pm$ 0.013} & \textbf{0.115 {\scriptsize $\pm$ 0.009}} & 0.088 {\scriptsize $\pm$ 0.013} & 0.079 {\scriptsize $\pm$ 0.011} & 0.037 {\scriptsize $\pm$ 0.025} & 0.091 {\scriptsize $\pm$ 0.019} & \textbf{0.153 {\scriptsize $\pm$ 0.028}} \\
        CoGymTravelProcess (agentRating) & \textbf{0.208 {\scriptsize $\pm$ 0.027}} & 0.185 {\scriptsize $\pm$ 0.061} & 0.115 {\scriptsize $\pm$ 0.050} & 0.101 {\scriptsize $\pm$ 0.044} & 0.121 {\scriptsize $\pm$ 0.056} & \textbf{0.218 {\scriptsize $\pm$ 0.246}} & 0.144 {\scriptsize $\pm$ 0.098} & 0.090 {\scriptsize $\pm$ 0.046} \\
        CoGymTravelProcess (communicationRating) & 0.172 {\scriptsize $\pm$ 0.075} & \textbf{0.285 {\scriptsize $\pm$ 0.066}} & 0.168 {\scriptsize $\pm$ 0.098} & 0.167 {\scriptsize $\pm$ 0.082} & 0.165 {\scriptsize $\pm$ 0.028} & \textbf{0.238 {\scriptsize $\pm$ 0.281}} & 0.220 {\scriptsize $\pm$ 0.154} & 0.180 {\scriptsize $\pm$ 0.195} \\
        CoGymTravelOutcome (outcomeRating) & 0.337 {\scriptsize $\pm$ 0.059} & 0.318 {\scriptsize $\pm$ 0.100} & 0.429 {\scriptsize $\pm$ 0.068} & \textbf{0.448 {\scriptsize $\pm$ 0.117}} & 0.413 {\scriptsize $\pm$ 0.057} & 0.298 {\scriptsize $\pm$ 0.472} & \textbf{0.558 {\scriptsize $\pm$ 0.131}} & 0.518 {\scriptsize $\pm$ 0.273} \\
        CoGymTabularProcess (agentRating) & 0.254 {\scriptsize $\pm$ 0.150} & 0.487 {\scriptsize $\pm$ 0.132} & 0.538 {\scriptsize $\pm$ 0.067} & \textbf{0.598 {\scriptsize $\pm$ 0.082}} & 0.403 {\scriptsize $\pm$ 0.108} & 0.475 {\scriptsize $\pm$ 0.203} & 0.560 {\scriptsize $\pm$ 0.395} & \textbf{0.637 {\scriptsize $\pm$ 0.315}} \\
        CoGymTabularProcess (communicationRating) & 0.360 {\scriptsize $\pm$ 0.046} & 0.608 {\scriptsize $\pm$ 0.088} & 0.791 {\scriptsize $\pm$ 0.093} & 0.779 {\scriptsize $\pm$ 0.147} & \textbf{0.798 {\scriptsize $\pm$ 0.049}} & --- & \textbf{0.890 {\scriptsize $\pm$ 0.125}} & 0.787 {\scriptsize $\pm$ 0.501} \\
        CoGymTabularOutcome (outcomeRating) & 0.363 {\scriptsize $\pm$ 0.217} & 0.349 {\scriptsize $\pm$ 0.173} & 0.367 {\scriptsize $\pm$ 0.160} & 0.201 {\scriptsize $\pm$ 0.081} & \textbf{0.634 {\scriptsize $\pm$ 0.117}} & --- & \textbf{0.363 {\scriptsize $\pm$ 0.362}} & 0.200 {\scriptsize $\pm$ 0.227} \\
        \midrule
        \textbf{Average} & 0.159 & 0.206 & 0.281 & 0.263 & 0.276 & 0.108 & 0.323 & 0.287 \\
    \bottomrule
  \end{tabular}%
  }
  \caption{Metric generation performance (Kendall's Tau) with 95\% confidence intervals over 5 independent runs. Each generator produces metrics using persistent train sets, then correlation with human annotations is measured on persistent validation sets. Cheap methods (left) generate 10 metrics per trial, expensive methods (right) generate 1 metric per trial (except finetune which generates 10). Results show correlation between generated metrics and ground-truth human annotations across diverse tasks using the Qwen3 32B model.}
  \label{tab:metric_gen_kendall_qwen3_32b}
\end{table*}

\begin{table*}[h]
  \centering
  \small
  \setlength{\tabcolsep}{6pt}
  \renewcommand{\arraystretch}{1.1}
  \resizebox{\textwidth}{!}{%
  \begin{tabular}{lccccc|ccc}
    \toprule
    \rowcolor{gray!30}
    & \multicolumn{5}{c|}{\textbf{Cheap to Produce}} & \multicolumn{3}{c}{\textbf{Expensive to Produce}} \\
    \rowcolor{gray!30}
    Task (Measure) & Code Gen & G-Eval & Single Criterion & Rubric (DSPy) & Rubric (Prometheus) & Finetune & Examples & MIPROv2 \\
    \midrule
        \rowcolor{gray!10}
        \multicolumn{9}{l}{\textit{\textbf{In-Distribution Tasks}: some metrics in our bank were designed to directly evaluate these tasks.}} \\
        \midrule
        SimpEval (score) & 0.127 {\scriptsize $\pm$ 0.015} & 0.279 {\scriptsize $\pm$ 0.024} & \textbf{0.324 {\scriptsize $\pm$ 0.026}} & 0.299 {\scriptsize $\pm$ 0.024} & 0.166 {\scriptsize $\pm$ 0.022} & 0.046 {\scriptsize $\pm$ 0.037} & 0.297 {\scriptsize $\pm$ 0.041} & \textbf{0.318 {\scriptsize $\pm$ 0.039}} \\
        SimpDA (fluency) & 0.135 {\scriptsize $\pm$ 0.020} & 0.534 {\scriptsize $\pm$ 0.016} & 0.510 {\scriptsize $\pm$ 0.014} & \textbf{0.573 {\scriptsize $\pm$ 0.012}} & 0.460 {\scriptsize $\pm$ 0.013} & 0.050 {\scriptsize $\pm$ 0.051} & \textbf{0.639 {\scriptsize $\pm$ 0.028}} & 0.635 {\scriptsize $\pm$ 0.018} \\
        SimpDA (meaning) & 0.246 {\scriptsize $\pm$ 0.028} & 0.570 {\scriptsize $\pm$ 0.012} & 0.551 {\scriptsize $\pm$ 0.007} & \textbf{0.601 {\scriptsize $\pm$ 0.022}} & 0.538 {\scriptsize $\pm$ 0.014} & 0.055 {\scriptsize $\pm$ 0.038} & \textbf{0.686 {\scriptsize $\pm$ 0.039}} & 0.643 {\scriptsize $\pm$ 0.022} \\
        SimpDA (simplicity) & 0.092 {\scriptsize $\pm$ 0.045} & 0.500 {\scriptsize $\pm$ 0.016} & \textbf{0.540 {\scriptsize $\pm$ 0.005}} & 0.535 {\scriptsize $\pm$ 0.026} & 0.463 {\scriptsize $\pm$ 0.031} & 0.041 {\scriptsize $\pm$ 0.058} & 0.621 {\scriptsize $\pm$ 0.039} & \textbf{0.622 {\scriptsize $\pm$ 0.019}} \\
        \midrule
        \rowcolor{gray!10}
        \multicolumn{9}{l}{\textit{\textbf{Out-of-Distribution Tasks}: no metric is specifically designed for these -- tests generalization and metric generation.}} \\
        \midrule
        EvalGenProduct (grade) & 0.190 {\scriptsize $\pm$ 0.032} & 0.204 {\scriptsize $\pm$ 0.063} & \textbf{0.225 {\scriptsize $\pm$ 0.069}} & 0.175 {\scriptsize $\pm$ 0.087} & 0.189 {\scriptsize $\pm$ 0.080} & 0.210 {\scriptsize $\pm$ 0.236} & 0.121 {\scriptsize $\pm$ 0.040} & \textbf{0.248 {\scriptsize $\pm$ 0.069}} \\
        EvalGenMedical (grade) & 0.190 {\scriptsize $\pm$ 0.032} & 0.204 {\scriptsize $\pm$ 0.063} & \textbf{0.225 {\scriptsize $\pm$ 0.069}} & 0.175 {\scriptsize $\pm$ 0.087} & 0.189 {\scriptsize $\pm$ 0.080} & 0.210 {\scriptsize $\pm$ 0.236} & 0.121 {\scriptsize $\pm$ 0.040} & \textbf{0.248 {\scriptsize $\pm$ 0.069}} \\
        CoGymTravelProcess (agentRating) & \textbf{0.201 {\scriptsize $\pm$ 0.032}} & 0.113 {\scriptsize $\pm$ 0.037} & 0.092 {\scriptsize $\pm$ 0.027} & 0.173 {\scriptsize $\pm$ 0.041} & 0.127 {\scriptsize $\pm$ 0.054} & 0.218 {\scriptsize $\pm$ 0.246} & \textbf{0.223 {\scriptsize $\pm$ 0.108}} & 0.067 {\scriptsize $\pm$ 0.041} \\
        CoGymTravelProcess (communicationRating) & \textbf{0.232 {\scriptsize $\pm$ 0.074}} & 0.176 {\scriptsize $\pm$ 0.038} & 0.070 {\scriptsize $\pm$ 0.035} & 0.154 {\scriptsize $\pm$ 0.083} & 0.228 {\scriptsize $\pm$ 0.017} & 0.238 {\scriptsize $\pm$ 0.281} & \textbf{0.312 {\scriptsize $\pm$ 0.086}} & 0.000 {\scriptsize $\pm$ 0.000} \\
        CoGymTravelOutcome (outcomeRating) & 0.286 {\scriptsize $\pm$ 0.049} & 0.400 {\scriptsize $\pm$ 0.103} & 0.450 {\scriptsize $\pm$ 0.114} & \textbf{0.474 {\scriptsize $\pm$ 0.051}} & 0.412 {\scriptsize $\pm$ 0.092} & 0.298 {\scriptsize $\pm$ 0.472} & 0.502 {\scriptsize $\pm$ 0.024} & \textbf{0.513 {\scriptsize $\pm$ 0.009}} \\
        CoGymTabularProcess (agentRating) & 0.273 {\scriptsize $\pm$ 0.199} & 0.555 {\scriptsize $\pm$ 0.098} & \textbf{0.697 {\scriptsize $\pm$ 0.139}} & 0.555 {\scriptsize $\pm$ 0.064} & 0.435 {\scriptsize $\pm$ 0.120} & 0.475 {\scriptsize $\pm$ 0.203} & 0.600 {\scriptsize $\pm$ 0.000} & \textbf{0.659 {\scriptsize $\pm$ 0.220}} \\
        CoGymTabularProcess (communicationRating) & 0.404 {\scriptsize $\pm$ 0.138} & 0.853 {\scriptsize $\pm$ 0.091} & 0.859 {\scriptsize $\pm$ 0.093} & \textbf{0.881 {\scriptsize $\pm$ 0.020}} & 0.763 {\scriptsize $\pm$ 0.091} & --- & 0.000 {\scriptsize $\pm$ 0.000} & \textbf{0.890 {\scriptsize $\pm$ 0.125}} \\
        CoGymTabularOutcome (outcomeRating) & 0.470 {\scriptsize $\pm$ 0.227} & 0.547 {\scriptsize $\pm$ 0.159} & 0.565 {\scriptsize $\pm$ 0.090} & 0.616 {\scriptsize $\pm$ 0.120} & \textbf{0.804 {\scriptsize $\pm$ 0.094}} & --- & 0.430 {\scriptsize $\pm$ 0.268} & \textbf{0.623 {\scriptsize $\pm$ 0.329}} \\
        \midrule
        \textbf{Average} & 0.237 & 0.411 & 0.426 & 0.434 & 0.398 & 0.184 & 0.379 & 0.456 \\
    \bottomrule
  \end{tabular}%
  }
  \caption{Metric generation performance (Kendall's Tau) with 95\% confidence intervals over 5 independent runs. Each generator produces metrics using persistent train sets, then correlation with human annotations is measured on persistent validation sets. Cheap methods (left) generate 10 metrics per trial, expensive methods (right) generate 1 metric per trial (except finetune which generates 10). Results show correlation between generated metrics and ground-truth human annotations across diverse tasks using the GPT-4o Mini model.}
  \label{tab:metric_gen_kendall_gpt_4o_mini}
\end{table*}
\section{Additional Experiments}
\label{appendix:additional_experiments}

\subsection{Robustness for All Metrics}

\begin{figure*}[h]
    \centering
    \includegraphics[width=1.0\textwidth]{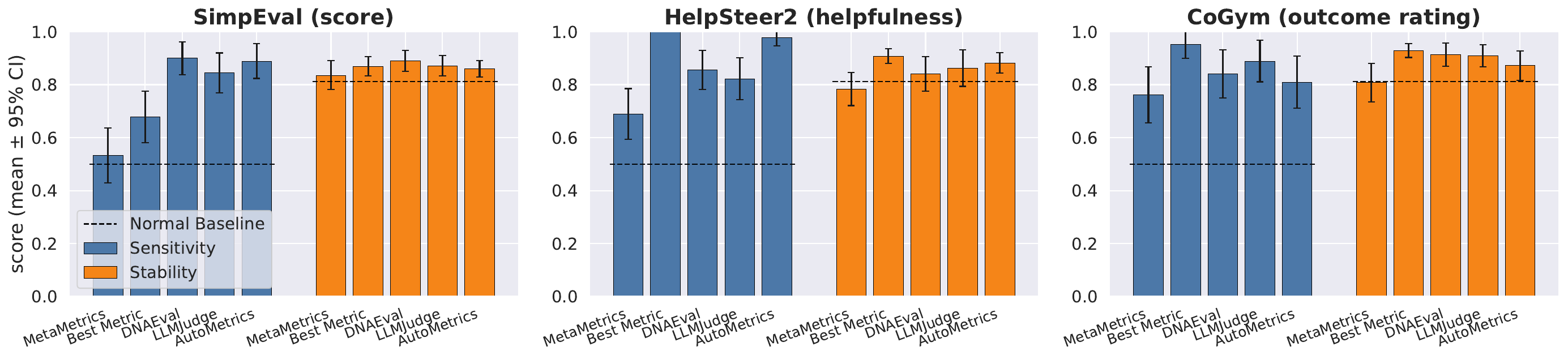}
    \caption{Sensitivity and Stability of all metrics for SimpEval, HelpSteer2, and CoGym.}
    \label{fig:sens_stab_all_metrics}
\end{figure*}

Here we include the results of the robustness experiment for all baseline metrics tested.  We report results in Figure~\ref{fig:sens_stab_all_metrics}.

We find that ``Best Metric" tends to be quite stable, while the LLM Based Metrics (DNAEval, LLMJudge, and AutoMetrics) stand out on robustness.

\subsection{What metrics does AutoMetrics actually select?}
\label{subsec:metric_exploration}

\begin{wrapfigure}{r}{0.45\textwidth}
    \vspace{-24pt}
    \includegraphics[width=\linewidth]{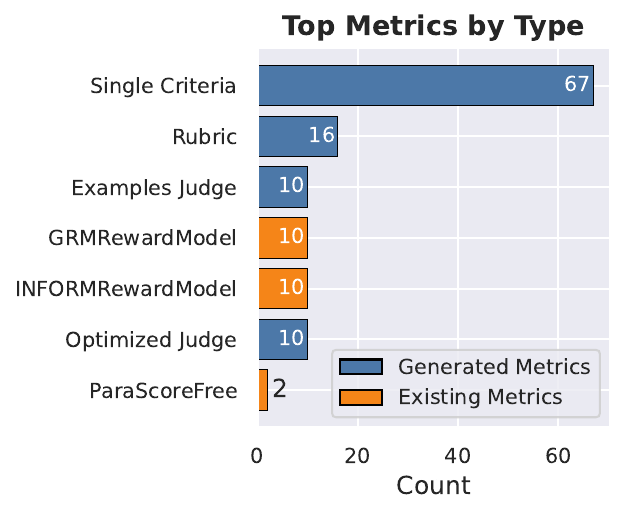}
    \caption{\small Breakdown of metrics recommended by AutoMetrics.  Generated are most common.}
    \label{fig:top_metrics}
    \vspace{-28pt}
\end{wrapfigure}

To explore the question of what metrics AutoMetrics actually recommends we turn to the 25 trials of AutoMetrics run for our main correlation experiment from Table~\ref{tab:main_results}. We look exclusively at the Qwen3-32B runs.  We provide a bar plot of metric types in Figure~\ref{fig:top_metrics}.

\paragraph{AutoMetrics are dominated by Generated Metrics.}  103 out of the 125 total recommended metrics were automatically generated.  Of the Existing metrics that were recommended 20 out of 22 were recommendations to use a reward model.  This suggests that the scope of metrics to retrieve from can be dramatically reduced to primarily recommending from the generated metrics as well as a few key reward models and other model based metrics like ``ParaScoreFree".  This insight will in practice greatly simplify the search space for metrics and lead to a more streamlined MetricBank.

\subsection{Validating Sensitivity and Stability}

In order to sanity check our sensitivity and stability scores we asked a collegue not involved in our project to annotate 150 datapoints from SimpEval \cite{maddela-etal-2023-lens} using the original annotation rubric described in the paper.  SimpEval consists of original and simple sentence pairs.  We asked them to annotate 30 pairs from the original dataset, 30 pairs where the simplified sentence was perturbed in a way that does not change the quality, and 90 sentences perturbed to purposefuly degrade the quality.  All perturbations were following our methodology described in \ref{subsec:evaluation-principles}.  Our human annotations yielded a sensitivity of 0.8275 and stability of 0.8000 suggesting the perturbations produced the intended effect.
\section{AutoMetrics Examples}
\label{sec:autometrics_examples}

\begin{AutoBox}[SimpEval --- score]

\textbf{Overall Kendall $\tau$:} 0.3234

\medskip
\textbf{Top 5 Metrics \& Coefficients}

\begin{tabular}{@{}L{0.70\linewidth} R{0.25\linewidth}@{}}
\toprule
\textbf{Metric} & \textbf{Coefficient} \\
\midrule
\texttt{Audience\_Appropriateness\_Qwen3-32B} & 1.7066 \\
\texttt{Conciseness\_Qwen3-32B} & 1.6676 \\
\texttt{Readability\_Score\_Qwen3-32B} & 1.6622 \\
\texttt{Clarity\_and\_Readability\_Rubric} & 1.6345 \\
\texttt{ParaScoreFree} & $-1.6125$ \\
\bottomrule
\end{tabular}

\medskip

\begin{AutoDesc}[Description: \texttt{Audience\_Appropriateness\_Qwen3-32B}]
Tailors language and phrasing to suit a general audience with minimal prior knowledge of the topic.
\end{AutoDesc}

\begin{AutoDesc}[Description: \texttt{Conciseness\_Qwen3-32B}]
Eliminates redundant phrases, wordiness, or tangential details while maintaining the original intent.
\end{AutoDesc}

\begin{AutoDesc}[Description: \texttt{Readability\_Score\_Qwen3-32B}]
Measures the text's ease of reading using standardized metrics (e.g., Flesch-Kincaid Grade Level).
\end{AutoDesc}

\begin{AutoDesc}[Description: \texttt{Clarity\_and\_Readability\_Rubric}]
\begin{lstlisting}[basicstyle=\ttfamily\small,breaklines=true,columns=fullflexible,keepspaces=true]
| Score | Description |
|-------|-------------|
| 1 | - The text is difficult to understand due to overly complex sentence structures, passive voice, or ambiguous phrasing.
- Redundant or redundant information is included, hindering clarity.
- Sentences are excessively long or fragmented, making it hard to follow the main idea.
- Jargon or technical terms are retained without simplification.
- The output fails to restructure the original sentence for broader accessibility. |
\end{lstlisting}
\begin{lstlisting}[basicstyle=\ttfamily\small,breaklines=true,columns=fullflexible,keepspaces=true]
| 2 | - The text is somewhat clear but still contains occasional complex structures or passive voice.
- Some sentences are overly long or include minor redundancies.
- Ambiguity or unclear phrasing is present in parts of the output.
- Simplification is attempted but incomplete, leaving some original complexity intact.
- The main idea is generally understandable but requires effort to parse. |
\end{lstlisting}
\begin{lstlisting}[basicstyle=\ttfamily\small,breaklines=true,columns=fullflexible,keepspaces=true]
| 3 | - The text is mostly clear, with mostly active voice and straightforward phrasing.
- Sentences are concise and well-structured, though a few may retain slight complexity.
- Minor ambiguities or redundancies are present but do not significantly hinder understanding.
- Simplification is effective for the core message, though some details may remain dense.
- The output is accessible to a general audience with minimal effort. |
\end{lstlisting}
\begin{lstlisting}[basicstyle=\ttfamily\small,breaklines=true,columns=fullflexible,keepspaces=true]
| 4 | - The text is clear and uses active voice consistently, with minimal passive constructions.
- Sentences are concise, well-structured, and free of unnecessary complexity.
- Ambiguity is largely avoided, and phrasing is precise.
- Simplification is thorough, with original complexity reduced to enhance accessibility.
- The output is easy to understand for a broad audience, with only minor improvements possible. |
\end{lstlisting}
\begin{lstlisting}[basicstyle=\ttfamily\small,breaklines=true,columns=fullflexible,keepspaces=true]
| 5 | - The text is exceptionally clear, using active voice and simple, direct sentence structures.
- All phrasing is unambiguous, and sentences are optimized for readability.
- Redundancy and complexity are entirely eliminated, with the core message distilled to its essentials.
- Simplification is flawless, making the content immediately accessible to all audiences.
- The output exemplifies best practices in clarity and readability, with no room for improvement. |


\end{lstlisting}
\end{AutoDesc}

\begin{AutoDesc}[Description: \texttt{ParaScoreFree}]
ParaScoreFree is a reference-free evaluation metric designed for paraphrase generation. It evaluates candidate paraphrases based on semantic similarity to the input source while encouraging lexical diversity. ParaScoreFree outputs a scalar quality score that combines BERT-based semantic similarity and normalized edit distance, offering a balance between meaning preservation and surface-level rewriting. It enables paraphrase evaluation without the need for gold reference texts, making it suitable for low-resource or open-domain settings.
\end{AutoDesc}

\end{AutoBox}

\begin{AutoBox}[HelpSteer2 --- helpfulness]

\textbf{Overall Kendall $\tau$:} 0.3481

\medskip
\textbf{Top 5 Metrics \& Coefficients}

\begin{tabular}{@{}L{0.70\linewidth} R{0.25\linewidth}@{}}
\toprule
\textbf{Metric} & \textbf{Coefficient} \\
\midrule
\texttt{INFORMRewardModel} & 0.2046 \\
\texttt{HelpSteer2\_helpfulness\_Qwen3-32B\_optimized\_seed45} & 0.1853 \\
\texttt{GRMRewardModel} & 0.1697 \\
\texttt{helpfulness\_Qwen3-32B\_examples} & 0.1661 \\
\texttt{Accuracy\_and\_Correctness\_Qwen3-32B} & 0.1625 \\
\bottomrule
\end{tabular}

\medskip

\begin{AutoDesc}[Description: \texttt{INFORMRewardModel}]
The INFORM Reward Model 70B (INF-ORM-Llama3.1-70B) is a large-scale outcome reward model designed to evaluate the quality of generated conversational responses. It predicts scalar reward scores for response texts, supporting preference-based fine-grained evaluations without requiring a reference response. The model is finetuned from the Llama-3.1-70B-Instruct backbone using preference-labeled datasets, employing scaled Bradley-Terry loss to incorporate preference magnitudes.
\end{AutoDesc}

\begin{AutoDesc}[Description: \texttt{HelpSteer2\_helpfulness\_Qwen3-32B\_optimized\_seed45}]
\begin{lstlisting}[basicstyle=\ttfamily\small,breaklines=true,columns=fullflexible,keepspaces=true]
Given the task description, evaluation axis, input/output texts, and suggested score range, analyze the output text's alignment with the task and axis by:  
1. **Assessing factual accuracy**: Verify if claims in the output are correct and supported by the input/text domain knowledge.  
2. **Evaluating relevance**: Determine if the output addresses the user's intent directly, avoiding verbosity or tangential content.  
3. **Analyzing structure and clarity**: Check if explanations are concise, logically organized, and accessible to the target audience (e.g., non-experts).  
4. **Identifying gaps or errors**: Highlight missing key details, misinterpretations, or inaccuracies that reduce helpfulness.  
5. **Scoring**: Assign a numerical score within the suggested range, balancing the above factors.  
Use the conversation history and task description as guidance for context and expectations. Prioritize precision in reasoning and alignment with the evaluation axis.

*Showing 0 of 8 total examples.*

\end{lstlisting}
\end{AutoDesc}

\begin{AutoDesc}[Description: \texttt{GRMRewardModel}]
The GRMRewardModel is a general-purpose reward model designed to evaluate the quality and safety of LLM-generated outputs. It achieves high generalization performance by applying a novel regularization method on hidden states during supervised fine-tuning. GRMRewardModel is fine-tuned on the decontaminated Skywork/Skywork-Reward-Preference-80K-v0.2 dataset and achieves state-of-the-art results among models of comparable size (3B), even outperforming some 8B reward models and proprietary LLM judges on RewardBench.
\end{AutoDesc}

\begin{AutoDesc}[Description: \texttt{helpfulness\_Qwen3-32B\_examples}]
\begin{lstlisting}[basicstyle=\ttfamily\small,breaklines=true,columns=fullflexible,keepspaces=true]
| Input Text | Score |
|------------|-------|
| <Input (prompt): <Can you teach me semi-definite programming in simple language?> Output (response): <Can you teach me how to use a computer in simple language?>> | 0 |
| <Input (prompt): <Delve into the nuanced benefits of engaging in group projects instead of the solo endeavor of individual projects. Craft your insights in a well-organized format, employing distinct headings for each category. Populate each section with a thoughtful list, elucidating each approach's merits and drawbacks. This approach aims to enhance clarity and the discussion's overall academ... | 0 |
| <Input (prompt): <The misery of life never appears in a clearer light than when a thinking person has quite plainly seen with horror its hazards and uncertainties and the total darkness in which he lives; how he cannot find anything solid, secure, and beyond dispute on to which he can hold; when, as I say, after such thoughts he does not at once destroy an existence that is not one, but breathi... | 1 |

*Showing 3 of 10 total examples.*
\end{lstlisting}
\end{AutoDesc}

\begin{AutoDesc}[Description: \texttt{Accuracy\_and\_Correctness\_Qwen3-32B}]
The factual correctness and reliability of the information provided.
\end{AutoDesc}

\end{AutoBox}

\begin{AutoBox}[EvalGenProduct --- grade]

\textbf{Overall Kendall $\tau$:} 0.4178

\medskip
\textbf{Top 5 Metrics \& Coefficients}

\begin{tabular}{@{}L{0.70\linewidth} R{0.25\linewidth}@{}}
\toprule
\textbf{Metric} & \textbf{Coefficient} \\
\midrule
\texttt{Formatting\_Compliance\_Qwen3-32B} & 0.1144 \\
\texttt{grade\_Qwen3-32B\_examples} & 0.1022 \\
\texttt{Call\_to\_Action\_\_CTA\_\_Strength\_Qwen3-32B} & 0.0752 \\
\texttt{Customer\_Review\_Integration\_Rubric} & 0.0747 \\
\texttt{Avoidance\_of\_Weaknesses\_Qwen3-32B} & 0.0653 \\
\bottomrule
\end{tabular}

\medskip

\begin{AutoDesc}[Description: \texttt{Formatting\_Compliance\_Qwen3-32B}]
Good examples strictly follow Markdown structure (headers, bullet points). Bad examples include disallowed elements (links, markdown errors).
\end{AutoDesc}

\begin{AutoDesc}[Description: \texttt{grade\_Qwen3-32B\_examples}]
\begin{lstlisting}[basicstyle=\ttfamily\small,breaklines=true,columns=fullflexible,keepspaces=true]
| Input Text | Score |
|------------|-------|
| <Input (Prompt): <You are an expert copywriter. You need to write an e-commerce product description based on the product details and customer reviews. Your description should be SEO-optimized. It should use an active voice and include the product's features, benefits, unique selling points without overpromising, and a call to action for the buyer. Benefits describe how product features will wor... | 0 |
| <Input (Prompt): <You are an expert copywriter. You need to write an e-commerce product description based on the product details and customer reviews. Your description should be SEO-optimized. It should use an active voice and include the product's features, benefits, unique selling points without overpromising, and a call to action for the buyer. Benefits describe how product features will wor... | 0 |
| <Input (Prompt): <You are an expert copywriter. You need to write an e-commerce product description based on the product details and customer reviews. Your description should be SEO-optimized. It should use an active voice and include the product's features, benefits, unique selling points without overpromising, and a call to action for the buyer. Benefits describe how product features will wor... | 1 |

*Showing 3 of 4 total examples.*
\end{lstlisting}
\end{AutoDesc}

\begin{AutoDesc}[Description: \texttt{Call\_to\_Action\_\_CTA\_\_Strength\_Qwen3-32B}]
Good examples include urgent, benefit-driven CTAs (e.g., 'Order now for seasonal savings'), while bad examples have vague or missing CTAs.
\end{AutoDesc}

\begin{AutoDesc}[Description: \texttt{Customer\_Review\_Integration\_Rubric}]
\begin{lstlisting}[basicstyle=\ttfamily\small,breaklines=true,columns=fullflexible,keepspaces=true]
| Score | Description |
|-------|-------------|
| 1 | - **No customer reviews included** or all quotes are fabricated.
- Reviews are irrelevant to the product or its benefits.
- Over-cites testimonials (e.g., 5+ quotes) or includes negative feedback.
- Quotes are generic (e.g., "Great product!") without specific context. |
| 2 | - **Minimal or inconsistent use of customer reviews** (e.g., 1-2 quotes).
- Quotes are vague or lack specificity (e.g., "I love this product!").
- Reviews may include irrelevant details or fail to align with the product's features/benefits.
- No clear connection between testimonials and the product's unique selling points. |
| 3 | - **Moderate use of customer reviews** (e.g., 2-3 quotes).
- Some quotes are specific and relevant (e.g., "This product works well for dry skin").
- May include 1-2 generic or slightly over-cited testimonials.
- Reviews are integrated but do not strongly enhance the description's persuasiveness. |
| 4 | - **Effective use of 1-2 authentic, contextually relevant quotes**.
- Testimonials highlight specific benefits (e.g., "The lightweight formula makes it perfect for travel").
- Quotes are concise, avoid over-citing, and align with the product's features.
- Reviews are integrated naturally into the description without overwhelming the reader. |
| 5 | - **Excellent integration of 1-2 highly specific, authentic testimonials**.
- Quotes directly tie to the product's unique selling points (e.g., "The smudge-proof formula lasts all day").
- Reviews are concise, impactful, and enhance the description's credibility.
- No fabricated, irrelevant, or over-cited quotes; testimonials feel organic and persuasive. |


*Showing 3 of 4 total examples.*
\end{lstlisting}
\end{AutoDesc}

\begin{AutoDesc}[Description: \texttt{Avoidance\_of\_Weaknesses\_Qwen3-32B}]
Good examples omit product drawbacks. Bad examples inadvertently mention flaws (e.g., 'may clog pores') or use hedging language.
\end{AutoDesc}

\end{AutoBox}

\begin{AutoBox}[RealHumanEval --- accepted]

\textbf{Overall Kendall $\tau$:} 0.1487

\medskip
\textbf{Top 5 Metrics \& Coefficients}

\begin{tabular}{@{}L{0.70\linewidth} R{0.25\linewidth}@{}}
\toprule
\textbf{Metric} & \textbf{Coefficient} \\
\midrule
\texttt{GRMRewardModel} & 0.0325 \\
\texttt{INFORMRewardModel} & 0.0293 \\
\texttt{Code\_Readability\_Qwen3-32B} & 0.0283 \\
\texttt{RealHumanEval\_accepted\_Qwen3-32B\_optimized\_seed44} & 0.0234 \\
\texttt{Modularity\_and\_Reusability\_Qwen3-32B} & 0.0218 \\
\bottomrule
\end{tabular}

\medskip

\begin{AutoDesc}[Description: \texttt{GRMRewardModel}]
The GRMRewardModel is a general-purpose reward model designed to evaluate the quality and safety of LLM-generated outputs. It achieves high generalization performance by applying a novel regularization method on hidden states during supervised fine-tuning. GRMRewardModel is fine-tuned on the decontaminated Skywork/Skywork-Reward-Preference-80K-v0.2 dataset and achieves state-of-the-art results among models of comparable size (3B), even outperforming some 8B reward models and proprietary LLM judges on RewardBench.
\end{AutoDesc}

\begin{AutoDesc}[Description: \texttt{INFORMRewardModel}]
The INFORM Reward Model 70B (INF-ORM-Llama3.1-70B) is a large-scale outcome reward model designed to evaluate the quality of generated conversational responses. It predicts scalar reward scores for response texts, supporting preference-based fine-grained evaluations without requiring a reference response. The model is finetuned from the Llama-3.1-70B-Instruct backbone using preference-labeled datasets, employing scaled Bradley-Terry loss to incorporate preference magnitudes.
\end{AutoDesc}

\begin{AutoDesc}[Description: \texttt{Code\_Readability\_Qwen3-32B}]
Clarity of variable names, structure, and comments for maintainability.
\end{AutoDesc}

\begin{AutoDesc}[Description: \texttt{RealHumanEval\_accepted\_Qwen3-32B\_optimized\_seed44}]
\begin{lstlisting}[basicstyle=\ttfamily\small,breaklines=true,columns=fullflexible,keepspaces=true]
You are an expert Python code reviewer in a high-stakes software engineering environment where code correctness directly impacts mission-critical systems (e.g., financial transactions, medical devices, or autonomous vehicles). Your task is to evaluate the AI-generated code output for **absolute correctness** and **completeness** along the specified evaluation axis. A single error could lead to catastrophic failures. Analyze the code with extreme rigor, checking for:  
1. **Logical correctness** (does it solve the task as described?)  
2. **Syntax validity** (Python 3 compliance, no placeholders like 'xrange()' or 'raw\_input()')  
3. **Edge case handling** (negative numbers, empty inputs, etc.)  
4. **Mathematical/statistical rigor** (valid algorithms, no arbitrary values like 'b = 8')  
5. **Functionality** (working return statements, no stubs or incomplete logic).  
Assign a score between 0.0 and 1.0, where 0.0 means the code is non-functional or completely ignores the task, and 1.0 represents a flawless implementation. Use the input/output text and conversation history for context.

*Showing 0 of 8 total examples.*
\end{lstlisting}
\end{AutoDesc}

\begin{AutoDesc}[Description: \texttt{Modularity\_and\_Reusability\_Qwen3-32B}]
Code organization into reusable functions/methods with clear separation of concerns.
\end{AutoDesc}

\end{AutoBox}

\begin{AutoBox}[CoGymTravelOutcome --- outcomeRating]

\textbf{Overall Kendall $\tau$:} 0.4301

\medskip
\textbf{Top 5 Metrics \& Coefficients}

\begin{tabular}{@{}L{0.70\linewidth} R{0.25\linewidth}@{}}
\toprule
\textbf{Metric} & \textbf{Coefficient} \\
\midrule
\texttt{Cultural\_and\_Local\_Integration\_Rubric} & 0.1963 \\
\texttt{Cultural\_and\_Local\_Experiences\_Qwen3-32B} & 0.1927 \\
\texttt{Accommodation\_Options\_Qwen3-32B} & 0.1824 \\
\texttt{outcomeRating\_Qwen3-32B\_examples} & 0.1674 \\
\texttt{Feasibility\_and\_Realism\_Qwen3-32B} & 0.1620 \\
\bottomrule
\end{tabular}

\medskip

\begin{AutoDesc}[Description: \texttt{Cultural\_and\_Local\_Integration\_Rubric}]
\begin{lstlisting}[basicstyle=\ttfamily\small,breaklines=true,columns=fullflexible,keepspaces=true]
| Score | Description |
|-------|-------------|
| 1 | - **Score 1 (Poor):**
- No mention of unique local experiences or cultural highlights.
- No authentic food/dining recommendations.
- Generic or irrelevant suggestions (e.g., luxury dining for a budget trip).
- Fails to address the user's query or intent. |
| 2 | - **Score 2 (Weak):**
- Minimal mention of local experiences (e.g., 1-2 generic activities like "visiting a market").
- Vague food/dining suggestions (e.g., "try local cuisine" without specifics).
- Lacks integration of cultural or seasonal traditions (e.g., no mention of KFC for Christmas).
- Missing links or references to local resources. |
| 3 | - **Score 3 (Fair):**
- Includes 1-2 specific local experiences (e.g., visiting Jigokudani Monkey Park).
- Mentions 1-2 authentic food/dining options (e.g., "try miso ramen").
- Some cultural or seasonal references (e.g., "KFC is popular for Christmas").
- Limited use of links or resources to support recommendations. |
| 4 | - **Score 4 (Good):**
- Includes 3-4 unique local experiences (e.g., snow monkeys, winter illuminations, regional festivals).
- Highlights 2-3 specific, culturally significant food/dining options (e.g., "try KFC for Christmas," "visit a local ramen shop").
- Integrates cultural/seasonal traditions (e.g., "Christmas markets in Hokkaido").
- Provides 1-2 links to local events, businesses, or resources. |
| 5 | - **Score 5 (Excellent):**
- Includes 5+ unique, deeply integrated local experiences (e.g., snow monkeys, winter illuminations, regional festivals, and lesser-known gems).
- Highlights 3+ specific, culturally significant food/dining options with detailed descriptions (e.g., "try miso ramen at [specific shop]").
- Seamlessly integrates cultural/seasonal traditions (e.g., "KFC for Christmas, winter illuminations in Sapporo").
- Provides 3+ links to local events, businesses, or resources (e.g., [Sapporo White Illumination](url)).
- Tailors recommendations to the user's preferences and avoids irrelevant suggestions. |
\end{lstlisting}

\end{AutoDesc}

\begin{AutoDesc}[Description: \texttt{Cultural\_and\_Local\_Experiences\_Qwen3-32B}]
Incorporates authentic local traditions, cuisine, and unique regional activities.
\end{AutoDesc}

\begin{AutoDesc}[Description: \texttt{Accommodation\_Options\_Qwen3-32B}]
Recommends diverse lodging choices (e.g., luxury hotels, budget hostels, or ryokans).
\end{AutoDesc}

\begin{AutoDesc}[Description: \texttt{outcomeRating\_Qwen3-32B\_examples}]
\begin{lstlisting}[basicstyle=\ttfamily\small,breaklines=true,columns=fullflexible,keepspaces=true]
| Input Text | Score |
|------------|-------|
| <Input (query): <I want to go to Hong Kong and spend less money> Output (outcome): <**Luxury Entertainment**  **Fine Dining:**  *   Michelin-starred restaurants: Hong Kong boasts a wide array of Michelin-starred restaurants offering diverse cuisines. Examples include Caprice (French), Lung King Hee (Cantonese), and Amber (Contemporary). *   Private dining experiences: Many high-end restaurants offer private dining rooms for a more... | 1.0 |
| <Input (query): <Kauai> Output (outcome): <\# Kauai Itinerary: 5 Days of Scenery \& Bro Time (Grad Student Budget)  **Trip Overview:** This itinerary focuses on experiencing Kauai's natural beauty while keeping costs down. It's designed for a group of bros looking for a relaxing and scenic getaway in late March.  **Weather Considerations:** March is the tail end of the rainy season in Kauai. Expe... | 2.0 |
| <Input (query): <I will go to yangzhou city in China> Output (outcome): <Thank you for sharing your interests in history and food! Could you please let me know how many days you plan to stay in Yangzhou? This will help me create a well-structured itinerary for your trip.>> | 2.0 |

*Showing 3 of 9 total examples.*
\end{lstlisting}
\end{AutoDesc}

\begin{AutoDesc}[Description: \texttt{Feasibility\_and\_Realism\_Qwen3-32B}]
Ensures the itinerary is realistic in terms of time, distance, and activity density.
\end{AutoDesc}

\end{AutoBox}

\begin{AutoBox}[TauBench --- reward]

\textbf{Overall Kendall $\tau$:} 0.2469

\medskip
\textbf{Top 3 Metrics \& Coefficients}

\begin{tabular}{@{}L{0.70\linewidth} R{0.25\linewidth}@{}}
\toprule
\textbf{Metric} & \textbf{Coefficient} \\
\midrule
\texttt{Membership\_Benefit\_Application\_Rubric} & 0.0803 \\
\texttt{Escalation\_Appropriateness\_Rubric} & 0.0599 \\
\texttt{Policy\_Compliance\_Qwen3-32B} & 0.0567 \\
\bottomrule
\end{tabular}

\medskip

\begin{AutoDesc}[Description: \texttt{Membership\_Benefit\_Application\_Rubric}]
\begin{lstlisting}[basicstyle=\ttfamily\small,breaklines=true,columns=fullflexible,keepspaces=true]
| Score | Description |
|-------|-------------|
| 1 | - **Score 1 (Fails to apply rules)**:
- Incorrectly assigns free baggage allowances regardless of membership tier or cabin class.
- Applies insurance benefits to users who do not meet eligibility criteria (e.g., no insurance, basic economy).
- Offers compensation certificates to users who are not eligible (e.g., regular members without insurance).
- Fails to enforce policy restrictions (e.g., allowing basic economy cancellations outside the 24-hour window without insurance). |
| 2 | - **Score 2 (Major errors in application)**:
- Applies baggage allowances inconsistently (e.g., correct for some tiers but not others).
- Misapplies insurance eligibility (e.g., allows refunds for cancellations without valid reasons).
- Offers compensation certificates in most cases but misses key eligibility criteria (e.g., ignores membership tier).
- Occasionally transfers to human agents unnecessarily due to incorrect benefit application. |
| 3 | - **Score 3 (Partial adherence with minor errors)**:
- Correctly applies baggage allowances for most membership tiers but has occasional errors (e.g., miscalculates free bags for gold members).
- Applies insurance eligibility in most cases but fails in edge cases (e.g., business class cancellations without checking insurance status).
- Offers compensation certificates in most eligible scenarios but occasionally misses conditions (e.g., delayed flights without verifying membership).
- Rarely transfers to human agents due to minor benefit application issues. |
| 4 | - **Score 4 (High adherence with rare errors)**:
- Correctly applies baggage allowances for all membership tiers and cabin classes in most cases.
- Applies insurance eligibility accurately in nearly all scenarios.
- Offers compensation certificates in all eligible cases but has one minor oversight (e.g., miscalculating certificate amounts for multi-passenger reservations).
- Transfers to human agents only when necessary and for valid reasons. |
| 5 | - **Score 5 (Perfect adherence)**:
- Always assigns free baggage allowances correctly based on membership tier and cabin class.
- Applies insurance eligibility and compensation rules flawlessly, adhering strictly to policy.
- Never offers ineligible benefits (e.g., no certificates to regular members without insurance).
- Transfers to human agents only when the request falls outside the scope of membership benefits. |

\end{lstlisting}
\end{AutoDesc}

\begin{AutoDesc}[Description: \texttt{Escalation\_Appropriateness\_Rubric}]
\begin{lstlisting}[basicstyle=\ttfamily\small,breaklines=true,columns=fullflexible,keepspaces=true]
| Score | Description |
|-------|-------------|
| 1 | - **Fails to transfer** in all cases where policy limits are reached or exceptions are needed.
- **Incorrectly handles** requests that require human intervention (e.g., proceeds with booking/canceling flights outside policy).
- **No adherence** to the rule of transferring for policy violations or exceptions. |
| 2 | - **Transfers inconsistently** (e.g., transfers in some policy-violating cases but not others).
- **Fails to transfer** for critical exceptions (e.g., basic economy cancellations without insurance, destination changes).
- **Attempts to resolve** issues beyond its scope (e.g., modifying flight destinations, waiving fees without human input). |
| 3 | - **Transfers** in most policy-violating cases (e.g., denies basic economy cancellations and transfers to human agents).
- **Partially handles exceptions** (e.g., transfers for compensation requests but not for all policy violations).
- **Some errors** in determining when to escalate (e.g., transfers unnecessarily for minor issues). |
| 4 | - **Consistently transfers** when policy limits are reached (e.g., denies basic economy cancellations, blocks destination changes).
- **Transfers for exceptions** (e.g., user insists on refunds for non-refundable tickets, requests compensation for delays).
- **Minimal errors** in escalation decisions, with clear adherence to policy boundaries. |
| 5 | - **Perfectly transfers** in all required cases (e.g., policy violations, exceptions, ambiguous requests).
- **Never attempts to handle** requests outside its scope (e.g., denies basic economy cancellations, blocks invalid modifications).
- **Proactively transfers** when user intent is unclear or requires human judgment (e.g., personal emergencies, compensation negotiations). |

\end{lstlisting}
\end{AutoDesc}

\begin{AutoDesc}[Description: \texttt{Policy\_Compliance\_Qwen3-32B}]
Adherence to airline rules (e.g., no basic economy cancellations without insurance or 24-hour window).
\end{AutoDesc}

\end{AutoBox}

\end{document}